\documentclass[authoryear, 5p, twocolumn, 11pt]{elsarticle}
\usepackage{amsfonts, amsmath, bm, amsthm, amssymb}
\usepackage{graphicx}
\usepackage{subfigure}
\usepackage{parskip} % remove indent and set space among paragraphs
\usepackage{setspace}
\usepackage{booktabs, multirow, makecell}
\usepackage[font=scriptsize]{caption}

\newcommand{\ra}[1]{\renewcommand{\arraystretch}{#1}} %expand interline in tables
\newcommand{\w}[1]{\widetilde{#1}}
\newcommand{\indep}{\perp \!\!\! \perp}
\newcommand{\notindep}{\not\!\perp\!\!\!\perp}
\DeclareMathOperator{\pa}{\mathbf{pa}}

\theoremstyle{definition}

\usepackage{natbib}
\usepackage[font=scriptsize]{caption}
\usepackage[hyphens]{url} % to go to newline in long urls (e.g. in biblio)
\usepackage{hyperref}
\hypersetup{
    urlcolor  = black,
	colorlinks = true,
	citecolor = blue, % citation color
	linkcolor = blue % internal links color
}

\usepackage{tikz}
\usetikzlibrary{shapes, decorations, arrows, calc, arrows.meta, fit, positioning}
\tikzset{
    -Latex,auto,node distance =1 cm and 1 cm,semithick,
    state/.style ={ellipse, draw, minimum width = 0.7 cm},
    point/.style = {circle, draw, inner sep=0.04cm, fill},
    bidirected/.style={Latex-Latex,dashed},
    el/.style = {inner sep=2pt, align=left, sloped}
}

\makeatletter
\def\ps@pprintTitle{%
 \let\@oddhead\@empty
 \let\@evenhead\@empty
 \def\@oddfoot{}%
 \let\@evenfoot\@oddfoot}
\makeatother

\begin{document}

%\title{The Zoo of Fairness metrics in Machine Learning}
\title{A Clarification of the Nuances in the Fairness Metrics Landscape\tnoteref{t1}}
\author[1,2]{Alessandro~Castelnovo\fnref{dr}}
\ead{alessandro.castelnovo@intesasanpaolo.com}

\author[1]{Riccardo~Crupi\fnref{dr}}
\ead{riccardo.crupi@intesasanpaolo.com}

\author[1,2]{Greta~Greco\fnref{dr}}
\ead{greta.greco@intesasanpaolo.com}

\author[1]{Daniele~Regoli\fnref{dr}}
\ead{daniele.regoli@intesasanpaolo.com}

\author[1]{Ilaria~Giuseppina~Penco\fnref{dr}}
\ead{ilaria.penco@intesasanpaolo.com}

\author[1]{Andrea~Claudio~Cosentini\fnref{dr}}
\ead{andrea.cosentini@intesasanpaolo.com}

\address[1]{Data Science \& Artificial Intelligence, Intesa Sanpaolo S.p.A., Italy}

\address[2]{Dept. of Informatics, Systems and  Communication, Univ. of Milano Bicocca, Italy}

\tnotetext[t1]{\emph{The final version of this paper is published by Scientific~Reports: \textbf{Sci Rep} 12, 4209 (2022) \href{https://doi.org/10.1038/s41598-022-07939-1}{doi.org/10.1038/s41598-022-07939-1}.}}

\fntext[dr]{The views and opinions expressed are those of the authors and do not necessarily reflect the views of Intesa Sanpaolo, its affiliates or its employees.}

\begin{abstract}
    In recent years, the problem of addressing fairness in Machine Learning (ML) and automatic decision-making has attracted a lot of attention in the scientific communities dealing with Artificial Intelligence.   
    A plethora of different definitions of fairness in ML have been proposed, that consider different notions of what is a ``fair decision'' in situations impacting individuals in the population. 
    The precise differences, implications and ``orthogonality'' between these notions have not yet been fully analyzed in the literature. In this work, we try to make some order out of this zoo of definitions.
\end{abstract}

    
\maketitle

\section*{Introduction}

The issue of discrimination bias in Artificial Intelligence (AI), in particular in Machine Learning (ML), has gained a lot of momentum in the scientific community in the last few years~\citep{barocas-hardt-narayanan, oneto2020fairness, mitchell2021algorithmic, mehrabi2021survey}.
The establishment of dedicated conferences such as the ACM Fairness, Accountability and Transparency (FAccT) conference\footnote{Formerly ACM FAT*, grown out of the FAT/ML workshops held from 2014 to 2018.}, has fostered the emergence of a new community of researchers working on topics that are sometimes roughly collected under the label of Trustworthy AI~\citep{hleg2019ethics, thiebes2021trustworthy} (also Responsible AI or Ethical AI) --- topics such as fairness and bias in ML, and explainability of ``black-box'' ML models, that focus on possible risks and drawbacks of the use of ML at scale in automated decision making. 

Contextually, many research institutions, private companies and public sector organizations have started issuing and taking positions on ethical principles and guidelines for the responsible use of AI systems~\citep{jobin2019global}.
Indeed, besides the benefits, many ethical concerns have emerged in recent years with the widespread use of such systems: bias amplification, data privacy, lack of transparency, human oversight, accountability, etc.~\citep{royal2017machine, fjeld2020principled, campolo2017ai, floridi2018ai4people, floridi2019establishing, floridi2021translating, hleg2019ethics}. 
The European Commission has recently published a proposal for what is going to be the first attempt ever to insert AI systems and their use in a coherent legal framework~\citep{EUproposal}: the proposal explicitly refers to the risk of bias discrimination of AI systems.
Critics to the ``non-responsible'' employment of AI systems and to the possible harms for people and society have also reached a broader audience thanks to popular books, such as \citet{pasquale2015black, o2016weapons, eubanks2018automating} and documentaries, e.g. \citet{codedbias}, raising awareness on these topics.

It is now widely recognized that data-driven decision making is not \emph{per se} safe from producing \emph{unfair} or \emph{biased} decisions, either for amplifications of biases already present in the data they learn from, or for algorithmic inaccuracies~\citep{barocas2016big, angwin2016machine, o2016weapons, mehrabi2021survey, mulligan2019thing}.
The scientific literature addressing the problem of fairness in ML has focused in the recent years on two main aspects: 
\begin{itemize}
    \item[i)] how to measure and assess fairness (or, equivalently, bias), and
    \item[ii)] how to mitigate bias in models when necessary.
\end{itemize}
Regarding the first aspect, in the last few years an incredible number of definitions have been proposed, formalizing different perspectives from which to assess and monitor fairness in decision making processes. A popular tutorial presented at the Conference on Fairness, Accountability, and Transparency in 2018 was titled ``21 fairness definitions and their politics''~ \citep{narayanan2018translation}. The number has grown since then.
The proliferation of fairness definitions is not \emph{per se} an issue: it reflects the evidence that fairness is a multi-faceted concept, and concentrates on itself different meanings and nuances, in turn depending in complex ways on the specific situation considered. 
As it is often the case with moral and ethical issues, conflicts are present between different and typically equally reasonable positions~\citep{horty2003reasoning, brink1994moral, thomson1976killing}.  

Researches, in proposing definitions, have focused on different intuitive notions of ``unfair decisions'', often considered as the ones impacting people in different ways on the basis of some personal characteristics, such as gender, ethnicity, age, sexual or political or religious orientations, considered to be \emph{protected}, or \emph{sensitive}. Relationships and interdependence of these sensitive variables with other features useful for making decisions is one of the crucial points, that entangles in complex ways the ultimate aim of any model, i.e. making efficient decisions, and the desired goal of not allowing unfair discrimination to impact people's lives.

Fairness notions proposed in the literature are usually classified in broad areas, such as: definitions based on parity of statistical metrics across groups identified by different values in protected attributes (e.g. male and female individuals, or people in different age groups); definitions focusing on preventing different treatment for individuals considered similar with respect to a specific task; definitions advocating the necessity of finding and employing causality among variables in order to really disentangle unfair impacts on decisions.

These three broad classes can be further seen as the result of two important distinctions:
\begin{enumerate}
    \item observational vs. causality-based criteria;
    \item group (or statistical) vs. individual (or similarity-based) criteria. 
\end{enumerate}

Distinction 1. discriminates criteria that are purely based on observational distribution of the data from criteria that try to first unveil causal relationships among the variables at play (mainly through a mixture of domain knowledge and opportune inference techniques) and then assess fairness in the specific situation. 

Distinction 2. discriminates criteria that focus on equality of treatment among groups of people from criteria requiring equality of treatment among couples of similar individuals.

Regarding aspect ii), i.e. that of removing bias when present, several approaches have been introduced in order to provide bias mitigation in data-driven decisions. Roughly speaking, they fall into three families, depending on the specific point along the ML pipeline in which they operate: pre-processing methods, that try to remove bias directly from data; in-processing techniques, imposing fairness either as a constraint or as an additional loss term during training optimization; post-processing methods, working directly on the outcomes of the model.

The present work focuses on the first of these aspects, and in particular deals with the aforementioned fact that the number of different metrics for fairness introduced in the literature has boomed. The researcher or practitioner first approaching this facet of ML may easily feel confused and somehow lost in this zoo of definitions. 
These multiple definitions capture different aspects of the concept of fairness but, at the best of our knowledge, there is still no clear understanding of the overall landscape where these metrics live.
This work aims to take a step in the direction of analyzing the relationship among the metrics, and trying to put order in the fairness landscape. Table~\ref{tab:metrics} provide a rough schematic list of fairness metrics discussed throughout the paper.

As general references for the definition of fairness in ML we refer to the book \citet{barocas-hardt-narayanan} and to the papers \citet{verma2018fairness, mitchell2021algorithmic, oneto2020fairness, mehrabi2021survey, makhlouf2021applicability} for more compact surveys. \citet{chouldechova2018frontiers, chouldechova2020snapshot} provide a general view of the state of the art of fairness in ML.

In this work, we shall mainly deal with the most common problem in supervised ML, the binary classification with a single sensitive attribute, and we shall only make a brief reference to the additional subtleties arising when multiple sensitive attributes are present. Despite this huge simplifications, the landscape of fairness definitions is nevertheless extremely rich and complex.

In the literature, some attempts to survey existing fairness definitions in ML are present~\citep{mitchell2021algorithmic, verma2018fairness, mehrabi2021survey, barocas-hardt-narayanan}, but, to the best of our knowledge, none has focused solely on metrics trying to disentangle and analyze in depth the interdependence and relationships among them. The contribution of this work lies not so much in the exhaustiveness of the taxonomy as in the attempt to clarify the nuances in the fairness metrics landscape with respect to the twofold dimensions listed above, namely group vs. individual notions and observational vs. causality-based notions.  
In this work we focus mainly on general and qualitative description of the fairness metrics landscape: building upon rigorous definitions we want to highlight incompatibilities and links between apparently different concepts of fairness.

The rest of this paper is organized as follows: section~\ref{sec:bias} proposes an account of the most important sources of bias in ML models; section~\ref{sec:individual} introduces individual fairness; section~\ref{sec:group} is devoted to the most prolific set of definitions, namely the group (or statistical) metrics; causality-based criteria are discussed in section~\ref{sec:causality}.

\begin{table*}[ht!]
\caption{\textbf{Fairness metrics.} Qualitative schema of the most important fairness metrics discussed throughout the paper.}
\label{tab:metrics}
\centering
\ra{1.4}
\resizebox{.9\textwidth}{!}{
\begin{tabular}{@{}lcc|ccr@{}}
  \toprule
  
  & \phantom{abc} & \multicolumn{2}{c}{\bf notion} & \textbf{use of $Y$} & \textbf{condition} \\
  \cmidrule{3-6}\\[-1ex]
  
  % group
  \multirow{5}{*}{\makecell[l]{group\\fairness}} && \multicolumn{2}{c}{Demographic Parity} & - & equal acceptance rate across groups\\
  && \multicolumn{2}{c}{Conditional Demographic Parity} & -$^*$ & \makecell[r]{equal acceptance rate across groups\\ in any strata}\\
  && \multirow{3}{*}{error parity} & Equal Accuracy & \checkmark & equal accuracy across groups\\
  && & Equality of Odds & \checkmark & equal FPR and FNR across groups\\
  && & Predictive Parity & \checkmark & equal precision across groups\\

  \cmidrule{3-6}\\[-1ex]
  
  % individual
  \multirow{2}{*}{\makecell[l]{individual\\fairness}} && \multicolumn{2}{c}{FTU/Blindness} & - & no explicit use of sensitive attributes\\
  && \multicolumn{2}{c}{Fairness Through Awareness} & -$^*$ & similar people are given similar decisions\\

  \cmidrule{3-6}\\[-1ex]
  
  % casuality
  \multirow{2}{*}{\makecell[l]{causality-based\\fairness}} && \multicolumn{2}{c}{Counterfactual Fairness} & - & \makecell[r]{an individual would have been given\\ the same decision if she had had different\\ values in sensitive attributes}\\
  && \multicolumn{2}{c}{\makecell[c]{path-specific\\ Counterfactual Fairness}} & - & \makecell[r]{same as above, but keeping fixed\\ some specific attributes}\\
 
  \bottomrule
  \multicolumn{6}{c}{\footnotesize\makecell[l]{$^*$ there are exceptions to these cases where $Y$ is actually employed, e.g. CDP conditioning on $Y$ becomes Equality of Odds,\\ and there are notions of individual fairness that use a similarity metric defined on the target space~\citep{berk2017convex}.}}

\end{tabular}
}
\end{table*}

\section{The problem of bias in data-driven decisions\label{sec:bias}}

Most sources of bias in data-driven decision making lie in the data itself and in the way in which they are collected. It is out of the scope of this manuscript to list and discuss all the possible shades of biases that can hide in the data~\citep{friedman1996bias, crawford2013hidden, mehrabi2021survey, baeza2018bias, hardt2014big}, we want nevertheless to say something about the main cases that can be encountered in many real-life scenarios.

The most important qualitative distinction, that we draw from \citet{mitchell2021algorithmic, barocas2016big, mehrabi2021survey}, lies between \emph{statistical} or \emph{representation biases}, i.e. when the model learns from biased sampling from the population, and \emph{historical} or \emph{societal biases}, i.e. when the model learns from data where decisions are already impacted by past prejudice.

\subsection{Statistical bias}

Generally speaking, \emph{statistical bias} occurs whenever the data used for model training are not representative of the true population. 
This can be due to a form of \emph{selection bias}, i.e. when the individuals appearing in the data come from a non-random selection of the full population. This happens, for example, in the context of credit lending, where the information of the repayment is known only for people that were granted the loan.

Another way in which statistical bias can enter the data is via \emph{systematic measurement errors}. This happens when the record of past errors and performance is systematically distorted, especially in the case of different amount of distortion for different groups of people.

Similarly, it may happen that data are systematically missing or poorly recorded for entire strata of the population.

One aspect of paramount importance, but often omitted, is the following: in the context of fairness, \emph{even in samples perfectly representative of the population} it may be that some protected group happens to be a minority group. All in all, minorities exist. In this case, especially when the minority group is highly unbalanced, a ML model can learn less accurately or may learn to discard errors on that group simply because it is less important in terms of crude counting. Again, notice that this is not a form of statistical bias \emph{tout court}, since in this case the ``under-representativeness'' of the protected group is a true feature of the population. Nevertheless, this is something very common, and one of the most important source of unfair discrimination against protected groups that happen to be minorities~\citep{hardt2014big}. 

This last issue may be reasonably considered as a source of biased decisions due to poor modeling rather than to a problem in data. The boundary is subtle, and in any case this is a matter of nomenclature: the important thing is to be aware of the risk.

\subsection{Historical or societal bias}

Even when the data are free from statistical bias, i.e. they truly represent the population, take into account minorities and there is no systematic error in recording, still it may be that bias exists simply because data reflect biased decisions.  

In most cases, this is due to a form of \emph{labelling bias}, i.e. a systematic favour/disfavour towards groups of people at the moment of creating the target variable from which the model is going to learn.
If the recorded outcomes are somehow due to human decisions (e.g. a model for granting loans may be trained on loan officers' past decisions), then we cannot in general trust their objectiveness and ``fairness''. 

Other forms of historical bias may be even more radical: gender bias has a rather long history,
and is embedded in all sorts of characteristics and features in such a way that it is difficult or even impossible to evaluate its impact and disentangle its dependence on other variables. Think for example of income or profession disparities, just to name a few out of many. 
Thus, this is a situation in which long-lasting biases cause systematic differences in features pertaining different groups of people.
Again, this is not a form of un-representativeness of the sample, it is a bias present in the full population.

Finally, bias can lurk in the data in a variety of additional ways~\citep{barocas2016big, barocas-hardt-narayanan, mehrabi2021survey, hardt2014big}, e.g. a poor selection of features may result in a loss of important information in disproportionate ways across groups~\citep{barocas2016big} --- among many others.

\section{Individual fairness\label{sec:individual}}

\subsection{Toy examples and notation}
To be as clear as possible, in what follows we shall make several references to 2 toy examples: credit lending decisions, and job recruiting. We call $A$ the categorical random variable representing the protected attribute, that for reference we shall take to be gender, we label with $X$ all the other (non-sensitive) random variables that the decision-maker is going to use to provide its yes/no decisions $\hat{Y} = f(X, A) \in \{0, 1\}$; while we label $Y\in \{0, 1\}$ the ground truth target variable that needs to be estimated/predicted -- typically by minimizing some loss function $\mathcal{L}(Y, \hat{Y})$. $\widetilde{X} = (X, A)$ collectively represent all the features of the problem.
We denote with lowercase letters the specific realizations of random variables, e.g. $\left\{(x_1, y_1), \ldots, (x_n, y_n)\right\}$ represent a dataset of $n$ independent realizations of $(X, Y)$.
We employ calligraphic symbols to refer to domain spaces, namely $\mathcal{X}$ denotes the space where features $X$ live\footnote{Technically, since we employ uppercase letters to denote random variables, it would be more proper to say that $\mathcal{X}$ is the image space of $X: \Omega \rightarrow \mathcal{X}$, where $\Omega$ is the event space, and that $x \in \mathcal{X}$. We shall nevertheless use this slight abuse of notation for sake of simplicity.}.

\subsection{Similarity-based criteria}
Individual fairness is embodied in the following principle: \emph{similar individuals should be given similar decisions}. This principle deals with the comparison of single individuals rather than focusing on groups of people sharing some characteristics.
On the other hand, group fairness starts from the idea that there are \emph{groups of people} potentially suffering biases and unfair decisions, and thus tries to reach equality of treatment for groups instead of individuals. 

The first attempt to deal with a form of individual fairness is in \citet{dwork2012fairness}, where the concept is introduced as a Lipschitz condition of the map $f$ from the feature space to the model space:
\begin{equation}
\label{eq:individual}    
    dist_Y(\hat{y}_i, \hat{y}_j) < L\times dist_{\widetilde{X}}(\widetilde{x}_i, \widetilde{x}_j),
\end{equation}
where $dist_Y$ and $dist_{\widetilde{X}}$ denote a suitable distance in target space and feature space, respectively, and $L$ is a constant. 
Loosely speaking, {\itshape a small distance in feature space (i.e. similar individuals) must correspond to a small distance in decision space (i.e. similar outcomes)}. They also propose an approach to enforce such kind of fairness by introducing a constrained optimization at training time, once a distance on the feature space is given.

The concept of individual fairness is straightforward, and it certainly resonates with our intuitive notion of ``equality''. Formula~\eqref{eq:individual} provides also an easy way to assess whether the decision model $\hat{Y} = f(\widetilde{X})$ satisfies such concept. However, the major drawback of this type of definitions lies in the subtle concept of ``similar individuals''.  Indeed, defining a suitable distance metric $dist_{\widetilde{X}}$ on feature space embodying the concept of similarity on ``ethical'' grounds alone is almost as tricky as defining fairness in the first place.

Indeed, while for target space the natural choice $dist_{Y}(\hat{Y}_i, \hat{Y}_j) = \rvert \hat{Y}_i - \hat{Y}_j\rvert$ will do, for feature space the natural choice of the euclidean distance on the space $\mathcal{A} \times \mathcal{X}$ implies that any smooth function $f$ shall satisfy condition~\eqref{eq:individual}, but this is not a very interesting notion of fairness.
The point is that one should come up with a distance that captures the essential features for determining that target, and that do not mix with sensitive attributes. But, again, this is not very different from defining what is fairness in a specific situation.  

Take, e.g., the job recruiting framework: what identifies the couples of individuals that should be considered similar and thus given the same chance of being recruited? Maybe the ones with same level of skills and experience irrespective of anything else?

One possibility is to define similar individuals as couples belonging to different groups with respect to sensitive features (e.g. male and female) but with the same values for all the other features. With this choice, what we are requiring is that its outcome should be unchanged if we take an observation and we only change its protected attribute $A$.
This concept is usually referred to as {\itshape Fairness Through Unawareness (FTU)} or {\itshape blindness} \citep{verma2018fairness}, and it is expressed as the \emph{requirement of not explicitly employing protected attributes in making decisions}.

Notice that this idea is very likely one of the first that one may think of when asked for, e.g., a decision-making process that does not discriminate against gender: not to explicitly use gender to make decisions, i.e. $\hat{Y} = f(X)$. Indeed, this concept is also referred to as \emph{disparate treatment}, i.e. there is disparate treatment whenever two individuals sharing the same values of non-sensitive features but differing on the sensitive ones are treated differently \citep{barocas2016big, zafar2017fairness}.

Unfortunately, despite its compelling simplicity, fairness through unawareness comes not without flaws.
Firstly, if it is true that reaching FTU is straightforward, it is not that easy to assess. The problem of \emph{bias assessment} in a model consists in measuring whether the model decisions are biased given a set of realizations of $(A, X, \hat{Y})$, namely $\left\{(a_1, x_1, \hat{y}_1), \ldots, (a_n, x_n, \hat{y}_n)\right\}$ -- the corresponding values of $Y$ are needed as well for some criteria. In this setting, it is tricky to measure whether FTU holds, the main reason being
that it is more a request on how the model works rather than a request on properties of the output decisions.
One possible candidate metric is the following:
\begin{equation}
\label{eq:consistency}
\text{consistency} = 1 - \frac{1}{n}\left(\sum_{i=1}^n \left\rvert \hat{y}_i - \frac1k\sum_{x_j \in kNN(x_i)} \hat{y}_j\right\rvert\right),
\end{equation}
introduced in \citet{zemel2013learning}. Namely, for each observation $(x_i, \hat{y}_i)$ it measures how much the decision $\hat{y}_i$ is close to the decisions given to its $k$ nearest neighbors $kNN(x_i)$ in the $\mathcal{X}$ space\footnote{Notice that also in the computation of $kNN$ one has to choose a distance function on $\mathcal{X}$.}. Notice that it may happen that the $k$ neighbors of, say, a male individual are all males: in this case consistency~\eqref{eq:consistency} would in fact be equal to 1, but this does not prevent the model from explicitly using $A$ in making decisions.
Another possibility, partly inspired by \citet{berk2009role}, would be to compute
\begin{equation}
\frac{1}{n_1 n_2}\sum_{\substack{a_i=1,\\ a_j=0}} e^{-dist(x_i, x_j)} \rvert \hat{y}_i - \hat{y}_j\rvert,
\end{equation}
which measures the difference in decisions among men and women weighted by their similarity in feature space: the higher its value the higher the difference in treatment for couples of similar males and females.\footnote{The term $e^{-dist(x_i, x_j)}$ can be substituted with any measure of similarity of the points $x_i$ and $x_j$.}
On the other hand, it is much easier to assess FTU if we also have access to the model: we could create a synthetic dataset by flipping $A$\footnote{This holds if $A$ is binary. But it is easy to come up with generalizations to the multiclass case.}: $\left\{(a'_1, x_1), \ldots, (a'_n, x_n)\right\}$, feeding it to the model to get the corresponding outcomes $\hat{y}_1',\ldots, \hat{y}'_n$, and then compute the average $\tfrac1n\sum_{i = 1}^n \rvert \hat{y}_i - \hat{y}'_i\rvert$.

However, the main drawback of FTU is the following: it does not take into account the possible interdependence between $A$ and $X$. Other features may contain information on the sensitive attribute, thus explicitly removing the sensitive attribute is not sufficient to remove its information from the dataset. Namely, it may be that in the actual dataset there is a very low chance that a male and a female have similar values in all the (other) features, since gender is correlated with some of them. One may or may not decide that (some of) these correlations are legitimate\footnote{E.g. gender may be correlated with income, but, depending on the problem, one may decide that the use of income, even if correlated with gender, is not a source of unfair discrimination. In section~\ref{sec:causality} we will dwell more on this issue, namely that there may be some information correlated to the sensitive attribute but still considered ``fair''.}. This is one of the reasons why the definition of a issue-specific distance is crucial for a more refined notion of individual fairness.

One straightforward way to deal with correlations is to develop a model that is blind not only with respect to the sensitive attribute, but also to all the other variables with sufficiently high correlation with it. This is a method known as {\itshape suppression} \citep{kamiran2012data}. Apart from the obvious issue in defining a good threshold for correlation above which a predictor should be removed, this approach has the main drawback in the potentially huge loss of legitimate information that may reside in features correlated with the sensitive attribute.

One line of research has tried to solve this issue by ``cleaning'' the training dataset in order to remove both the sensitive attribute \emph{and} the information coming from the sensitive attribute ``lurking'' inside other features, while \emph{keeping the features}. This can be done, e.g., by using residuals of regressions of the (correlated) features on the sensitive attribute \citep{berk2009role, johndrow2019algorithm}, i.e. projecting the feature space in a subspace orthogonal to $A$. However, it is tricky to account for dependencies on feature interactions.
Another way of keeping information while removing ``unfairness'' due to the sensitive attribute is to learn a {\itshape fair representation} of the training dataset, see \citet{zemel2013learning, louizos2015variational, mcnamara2017provably, calmon2017optimized}, i.e. to learn a new set of variables $Z$, such that they are able to reconstruct $X$ with as few errors as possible, while at the same time being as independent as possible of $A$\footnote{In the sense that it is hard to reconstruct $A$ from $Z$.}.
The idea is then that one can use any ML model on this ``cleaned'' representation of the original dataset and this will be fair by design.

Notice, however, that in most situations, when trying to remove from the data the dependence of $A$ and $X$, individual fairness will be spoiled. We shall discuss this in section~\ref{sec:groupvsindividual}.

Counterfactual frameworks \citep{kusner2017counterfactual, chiappa2019path}, that will be more thoroughly discussed in section~\ref{sec:causality}, provide a clear way in which this similarity should be thought of: a male individual is similar to himself in the counterfactual world where he is a woman.      
Notice that this is crucially different from fairness through unawareness approach: a male individual transported in the counterfactual world where he is a woman will have differences in other features as well, and such differences are precisely due to the causal structure among the variables (very roughly speaking, this is the ``causal way'' to account for correlations). As an example, in the now popular Judea Pearl's assessment of the 1973 UC Berkeley admission case~\citep{bickel1975sex, pearl2016causal, pearl2018book} where there was different admission rate between men and women, being a woman 
``causes'' the choice of higher demanding departments, thus impacting the admission rate (in causal theory jargon, department choice is said to be a \emph{mediator} from gender to admission)~\citep{barocas-hardt-narayanan}. In this case, a male transported in the counterfactual world where he is a woman would have himself chosen higher demanding departments, thus this feature would change as well. In this case, a simple gender-blind model would not guarantee individual fairness. Indeed attribute flipping in general does not produce valid counterfactuals~\citep{black2020fliptest}. More details on this will be given in section~\ref{sec:causality}.

Turning to the general similarity-based definition~\eqref{eq:individual}, some work has been done to address the problem of finding a suitable similarity metrics in feature space. E.g. \citet{dwork2012fairness} and \citet{jung2019eliciting, ilvento2019metric} introduce the possibility to learn a issue-specific distance from data and from the contributions of domain experts. Indeed, the simple idea of using standard similarities, e.g. related to the euclidean distance on feature space, does not take into account the  trivial fact that some feature are more important than other in determining the relationship of an individual to specific target. Namely, for two applicants for a loan, the difference in income is much more important than the difference in, say, age, or even profession. Thus, judging what does it mean to be similar \emph{with respect to a specific task} is not that simple, and is in some sense connected also to the ground truth target variable $Y$.

Indeed, \citet{berk2017convex} propose a notion of individual fairness as a penalty function -- to be added to the risk minimization fitting -- which shifts the concept of similarity from the feature space to the target space: namely two individuals are deemed similar if they have a similar value of the target variable. Thus, this notion of individual fairness relies directly on the target attribute to define a task-specific distance. Of course, this definition is prone to biases possibly present in the target.

Broadly speaking, the Lipschitz notion \eqref{eq:individual} is sometimes referred to as Fairness Through Awareness (FTA), as opposed to Fairness Through Unawareness: in fact, even if they share the same principle of treating equally similar individuals, FTA is generally meant to use similarity metrics that are problem and target specific, i.e. that derive from an ``awareness'' of the possible impact, while FTU is a simple recipe that does not depend on the actual scenario.

\section{Group fairness}
\label{sec:group}

Group fairness criteria are typically expressed as conditional independence statements among the relevant variables in the problem. 

There are three main broad notions of observational group fairness, {\itshape independence}, {\itshape separation}, {\itshape sufficiency} \citep{barocas-hardt-narayanan}. Independence is strictly linked to what is known as {\itshape Demographic Parity} or {\itshape Statistical Parity}, separation is related to {\itshape Equality of Odds} and its relaxed versions, while sufficiency is connected to the concept of {\itshape calibration} and {\itshape Predictive Parity}.

There is a crucial aspect that discriminates independence criteria from the others: \emph{independence criteria rely only on the distribution of features and decisions, namely on $(A, X, \hat{Y})$, while separation and sufficiency criteria make use of the target variable $Y$ as well}.
This is an important thing to have in mind when trying to find your way in the zoo of fairness criteria in a specific case study:  when using separation-based or sufficiency-based criteria one must be careful to check whether the target variable $Y$ is itself prone to sources of bias and unfairness. 

In the example of credit lending, if $Y$ represents the decision of a loan officer, than separation-based and sufficiency-based criteria must be used with particular care, since $Y$ can be itself biased against some groups of people.

Actually, even if $Y$ stands for the repayment (or lack of it) of the loan, a form of \emph{selection bias} is very likely at work: we have that information only on applicants that received the loan in the first place, and these are almost surely not representative of the whole population of applicants.

\citet{zafar2017fairness} introduce the notion of \emph{disparate mistreatment} to refer to all group fairness criteria relying on disparity of errors, thus, in general, all group metrics dealing with comparisons among decisions $\hat{Y}$ and true outcomes $Y$.

Moreover, independence is said to be a \emph{non-conservative} measure of fairness\cite{raz2021group} --- meaning that it forces to change the \emph{status quo} --- since it is in general not satisfied by the perfect classifier $\hat{Y} = Y$ (apart from the trivial case when $Y\indep A$). Error rate parities, on the other hand, are \emph{conservative}, since they trivially hold for the perfect predictor.

\subsection{Independence\label{sec:independence}}

The criterion of independence~\citep{barocas-hardt-narayanan} states that the {\itshape decisions should be independent of any sensitive attribute}:
\begin{equation}
    \hat{Y} \indep A.
\end{equation}
This can be also expressed as follows:
\begin{equation}\label{eq:dp}
P(\hat{Y}=1\ \rvert\ A=a)=P(\hat{Y}=1\ \rvert\ A=b),\quad\forall a,b\in \mathcal{A},
\end{equation}
i.e. the ratio of loans granted to men should be equal to the ratio of loans granted to women.

The ratio of favorite outcomes is sometimes known as positive prediction ratio (ppr), thus independence is equivalent to requiring the same positive prediction ratio across groups identified by the sensitive features.
This form of independence is usually known as {\itshape Demographic Parity (DP)}, {\itshape statistical parity}, or sometimes as {\itshape group fairness}.\footnote{Demographic parity is a very common concept in the literature on fairness in ML. \citet{kamiran2009classifying} provides one of the first mathematical formulation (even if without actually using the now common nomenclature). We refer to \citet{barocas-hardt-narayanan} and \citet{chouldechova2017fair} for a general account.}

\begin{figure}
    \centering
    \includegraphics[width=.4\textwidth]{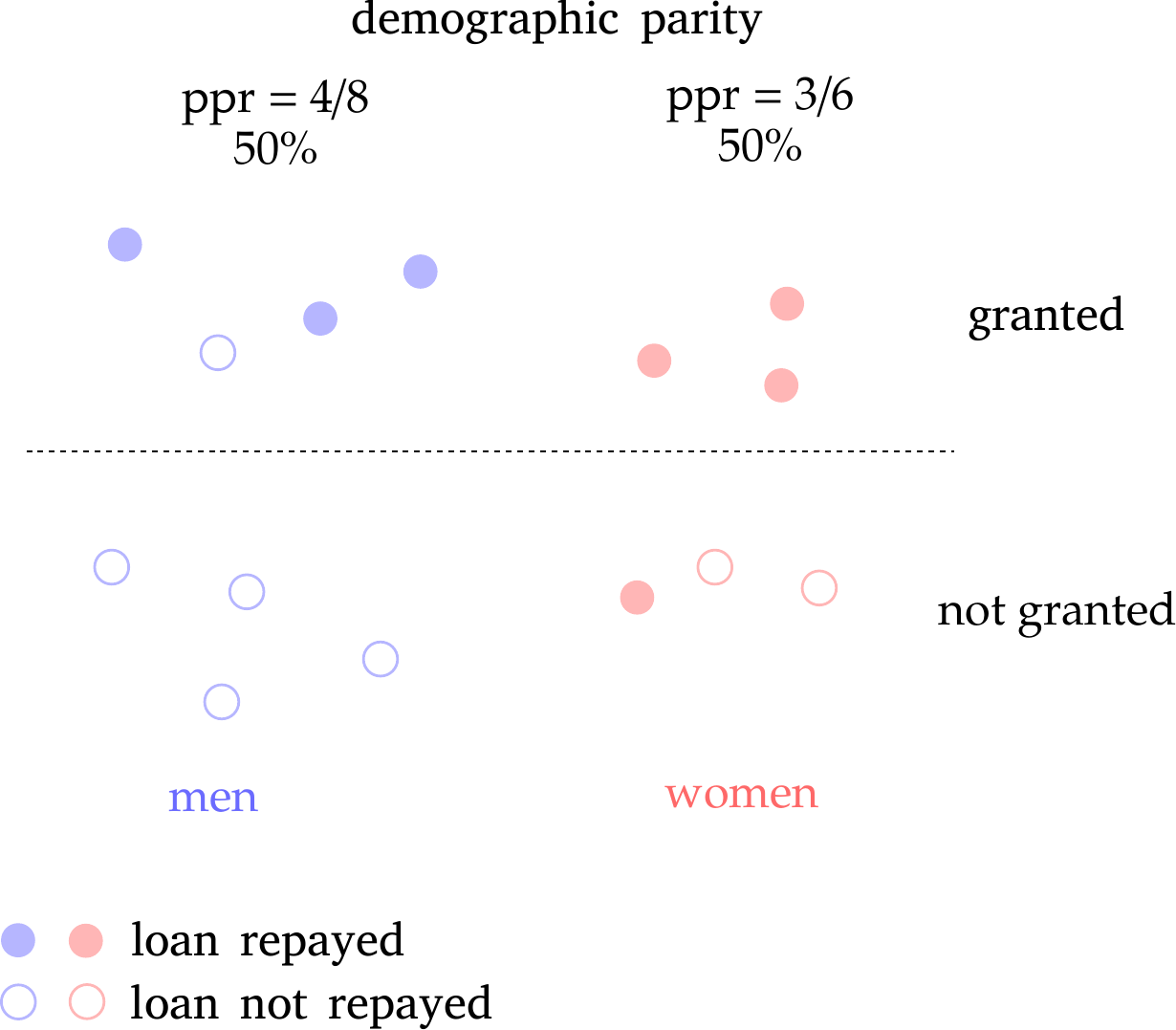}
    \caption{example of demographic parity in gender in credit lending toy model.}
    \label{fig:DP}
\end{figure}

Figure \ref{fig:DP} shows a very simple visualization of a model reaching demographic parity among men and women.

If some group has a significantly lower positive prediction ratio with respect to others, we say that demographic parity is not satisfied by the model decisions $\hat{Y}$.
In order to have a single number summarizing the amount of disparity, it is common to use either maximum possible difference or the minimum possible ratio of positive prediction ratios: a difference close to 0 or a ratio close to 1 indicate a decision system \emph{fair} with respect to $A$ in the sense of DP. Typically, some tolerance is considered by employing a threshold below (or above) which the decisions are still considered acceptable. 

In the literature, it is common to cite the ``$4/5$ rule'' or ``80\% rule'': the selection rate of any group should be not less than $4/5$ than the one of the group with the highest selection rate, i.e DP ratio greater than $80\%$. This is a reference to the guidelines by the US Equal Employment Opportunity Commission (EEOC)~\citep{eeoc, feldman2015certifying}, and it is often cited as one of the few examples of a legal framework relying on a specific definition of fairness, in particular on the notion of Demographic Parity. However, it is usually overlooked that the guidelines explicitly state that this should be considered only as a rule of thumb, with many possible exceptions, and in particular they prescribe that business or job-related necessities can justify a lower DP ratio, depending on the circumstances. This is actually a step away from pure DP towards more individual notions, such those of Conditional Demographic Parity and error parities, that shall be discussed in sections~\ref{sec:cdp}, \ref{sec:separation}, \ref{sec:sufficiency}.

\subsubsection{Subtleties of demographic parity}

The meaning of DP is intuitive only to a superficial analysis. For example, one may at first think that removing the sensitive attribute from the decision making process is enough to guarantee independence and thus demographic parity.

In general, this is not the case. Take the credit lending example and assume that, for whatever reason, women tend to actually pay back their loans with higher probability than men. 
If this is the case, it is reasonable to assume that a rating variable that we call $R$ will be higher for women than for men. In this scenario, the sensitive attribute gender ($A$) is correlated with rating.
If the model uses only rating (but not gender) to compute its decision, there will be a higher rate of loans granted to women than to men, resulting in a demographic {\itshape disparity} among these groups. In this case, if you want to reach DP, the model needs to \emph{favor men over women}, granting loans to men with a lower rating threshold with respect to women.

Thus, because of interdependence of $X$ and $A$, not only it is not enough to remove the sensitive attribute from the decision making process, but if you want to have DP you need, in general, to {\itshape treat different groups in different ways}, precisely in order to compensate for the (unwanted) effect of this dependence. This is somehow the opposite of an intuitive notion of fairness!

\begin{figure}
    \centering
    \includegraphics[width=.4\textwidth]{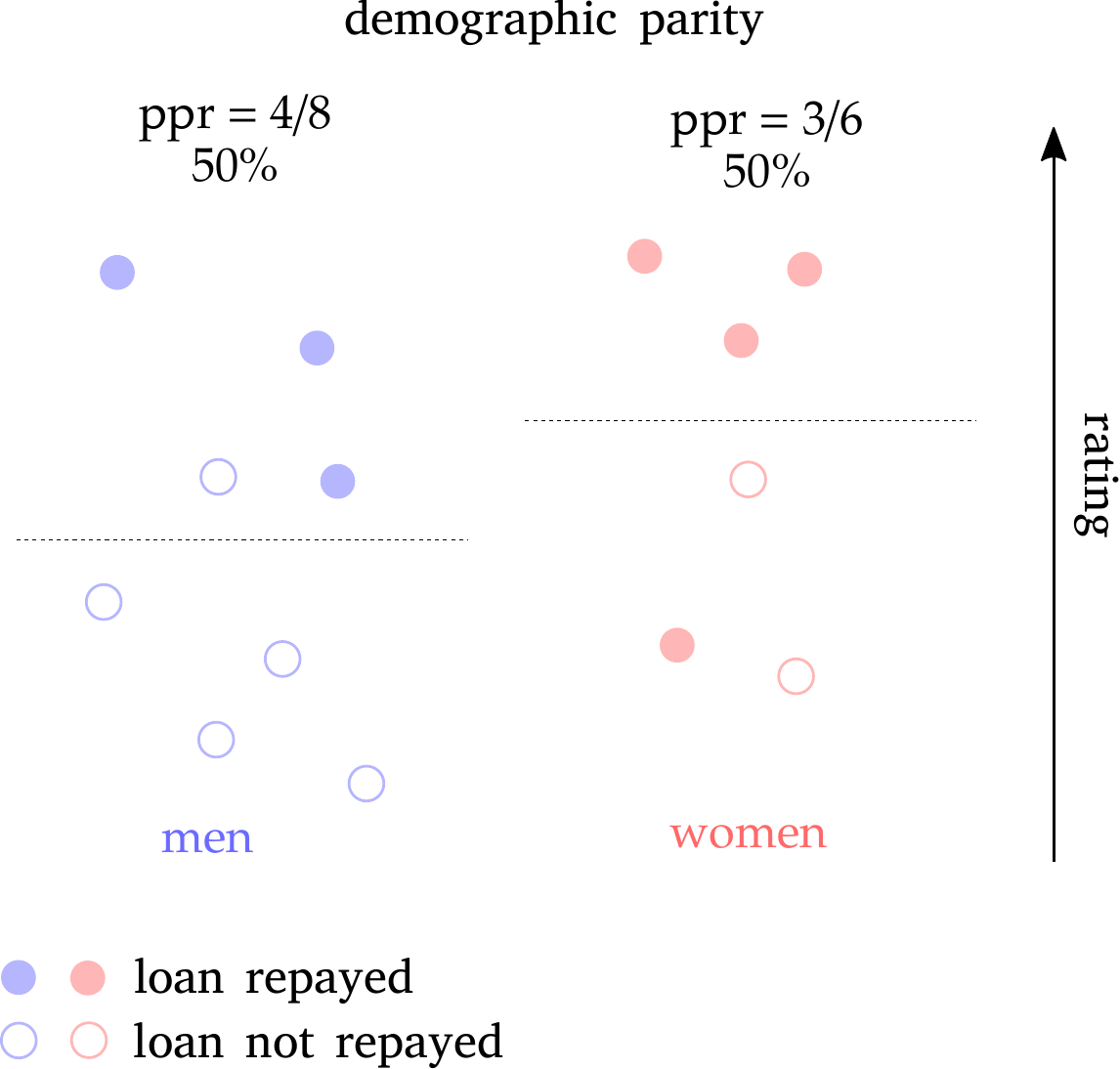}
    \caption{example of a subtlety of demographic parity: in order to reach demographic parity between men and women and still using rating as fundamental feature, one must use a different threshold between the groups, thus manifestly treating differently men and women.}
    \label{fig:DP_1}
\end{figure}

Figure \ref{fig:DP_1} displays this example: to reach equal ppr, one must use a different rating threshold for each group.

This, in turn, reveals another subtlety: even if your dataset and your setting does not apparently contain any sensitive features, discrimination could lurk in via correlations to sensitive features that you are not even collecting.

We want to stress here what we think is a crucial aspect: another possibility to reach DP would be to use \emph{neither gender nor rating in making decisions}, i.e. trying to remove all gender information from the dataset. Notice that this could be problematic for the accuracy of your decisions, since it's plausible that, by removing all variables correlated to $A$, information useful to estimate the target is lost as well. This is called \emph{suppression}, and was discussed already in section~\ref{sec:individual}, together with the concept of \emph{fair representation}, by which one tries to remove all sensitive information from the dataset while keeping as much useful information as possible. %As we shall argue in section~\ref{sec:groupvsindividual}, this approach allows to satisfy DP, while not favouring any subgroups with respect to $A$. 

Another possible flaw is the following: if it is true that women repay their loans with higher probability, is it really fair to have demographic parity between men and women? Why the bank should agree to grant loans with the same rate to groups that actually pays back with different probabilities? Should we stick to actual repayment rates or we should ask \emph{why} these probabilities are different and if this is possibly due to gender discrimination in the first place?

We try to summarize, in a non exhaustive list, a set of scenarios in which it might be reasonable to take into account demographic parity, among other metrics:
\begin{itemize}[-]
    \item when you want to actively {\itshape enforce} some form of equality between groups, irrespective of all other information. Indeed there are some  characteristics that are widely recognized to be independent \emph{in principle} of sensitive attributes, e.g. intelligence and talent, and there may be the need to enforce an independence in problems where the decisions is fundamentally linked to those characteristics; more in general, there may be reasons to consider unfair any relation among $A$ and $Y$, even if the data (objective data as well) is telling differently;
    \item (intertwined with the previous) when you deem that in your specific problem there are hidden historical biases that impact in a complex way the entire dataset;
    \item when you cannot trust the objectivity of the target variable $Y$, then demographic parity still makes sense, while many other metrics don't (e.g. separation).
\end{itemize}

Justifying the use of DP as a metric with the idea of enforcing an equality that should hold in principle has been discussed in the literature~\citep{friedler2016possibility, hertweck2021moral, raz2021group}. We would like to spend here a word of care in this respect, namely on the fact that favouring a group does indeed guarantee that the equality is reached in that specific use case, but it may trigger mechanisms that actually \emph{amplify} the bias that we intended to break. In the example of job recruiting, imposing the decision maker to increase the acceptance rate on a group of people with (relatively) low skill levels, will increase the likelihood that applicants hired from this group will perform poorly relative to the average, effectively perpetuating the bias in the way in which this group is perceived by others.  
In general, assessing the future impact of imposing a fairness condition, i.e. considering fairness as a dynamical process rather than a one-shot condition, is a subtle issue by itself~\citep{jabbari2017fairness, liu2018delayed, hu2018short}.

\subsubsection{Conditional demographic parity\label{sec:cdp}}

Another version of the independence criteria is the one of {\itshape Conditional Demographic Parity (CDP)}, first introduced in~\citet{kamiran2013quantifying}.
In the example given above, we may think that a fairer thing to do, with respect to full independence, is to require independence of the decision on gender only for men and women with the same level of rating. 
In other words, if a man and a woman have both a certain level of rating, we want them to have the same chance of getting the loan.
This goes somewhat in the direction of being an individual form of fairness requirement, since parity is assessed into smaller groups with respect to the entire sample.

Formally, this results in requiring {\itshape $\hat{Y}$ independent of $A$ given $R$}, \begin{equation}
    \hat{Y} \indep A\ \rvert\ R,    
\end{equation}
or, in other terms:
\begin{multline}\label{eq:cdp}
    P(\hat{Y}=1\ \rvert\ A=a,R=r)=P(\hat{Y}=1\ \rvert\ A=b,R=r),\\
    \forall a,b\in \mathcal{A},\forall r.
\end{multline}

This seems a very reasonable requirement in many real-life scenarios. For example, if you think of a recruitment setting where you don't want to bias women against men, but still you want to recruit the most skilled candidates, you may require your decision to be independent of gender but conditional on a score based on the curriculum and past work experiences: among people with an ``equivalent'' set of skills, you want to recruit men and women with the same rate.\\
In other words, the only disparities that you are willing to accept between male and female candidates are those justified by curriculum and experience.

Unfortunately, also in this case one must be really careful, because the variable that you are conditioning on \emph{might itself be a source of unfair discrimination}.  For example, it might well be that rating is higher for women not because they actually pay back their debts more likely than men, but because the rating system is biased against men.
And this may be due to self-fulfilling prophecy: if men have lower rating they may receive loans with higher interest rates, and thus have a higher probability of not paying them back, in a self-reinforcing loop.

Moreover, it may be not be so straightforward to select the variables to condition on: why condition on rating and not, e.g., on level of income, or profession, etc...?

Finally, in line with what outlined above for demographic parity, one may argue that women's curricula and work experiences tend to be on average different than those of male candidates for historical reasons and for a long-lasting (and die-hard) man-centered society. This may suggest that plain demographic parity could be more appropriately enforced in this case to reach a ``true'' equality.

Incidentally, notice that pushing to the extreme the notion of conditional demographic parity, i.e. conditioning on \emph{all the (non-sensitive) variables}, one has
\begin{equation}
\label{eq:cdp_strong1}
    \hat{Y} \indep A\ \rvert\ X,
\end{equation}
i.e.
\begin{multline}
\label{eq:cdp_strong2}
    P(\hat{Y} = 1\ \rvert\ A=a, X=x) =
    P(\hat{Y} = 1\ \rvert\ A=b, X=x),\\ \forall a, b \in \mathcal{A}, \forall x \in \mathcal{X};    
\end{multline}
meaning that a male and a female \emph{with the same value for all the other features} must be given the same outcome. This criterion is reachable by a gender-blind model, thus is strictly connected to the notion of FTU. We can indeed notice -- but leave the details to section~\ref{sec:groupvsindividual} -- that conditioning is equivalent to restrict the groups of people among which we require parity, thus is a way to go in the direction of obtaining an individual criterion.

\subsection{Separation \label{sec:separation}}

Independence and conditional independence do not make use of the true target $Y$.   
What if, instead of conditioning over rating $R$ we condition on the target $Y$? 

This is equivalent to requiring the independence of the decision $\hat{Y}$ and gender $A$ separately for individuals that actually repay their debt and for individuals that don't.
Namely, among people that repay their debt (or don't), we want to have the same rate of loan granting for men and women.

This concept has been called separation~\citep{barocas-hardt-narayanan}:
\begin{equation}
    \hat{Y} \indep A\ \rvert\ Y.
\end{equation}
In other terms
\begin{multline}
\label{eq:eo}
    P(\hat{Y}=1\ \rvert\ A=a,Y=y) = P(\hat{Y}=1\ \rvert\ A=b,Y=y),\\
    \forall a,b\in \mathcal{A},\ y\in\{0,1\}.
\end{multline}
Equivalently, {\itshape disparities in groups with different values of $A$} (male and female) {\itshape should be completely justified by the value of $Y$} (repayment or not).

As in the conditional independence case, this seems a very reasonable fairness requirement, {\itshape provided that you can completely trust the target variable}.  Namely, one should be extremely careful to check whether the target $Y$ is not itself a source of bias.

For example, if $Y$ instead of reflecting true repayment was the outcome of loan officers' decision on whether to grant the loans, it could incorporate bias, thus we it would be risky to assess fairness with direct comparisons with $Y$. Moreover, as said above, even in the objective case where $Y$ is the actual repayment, a form of selection bias would likely distort the rate of repayment.

We can express separation in terms of what are known in statistics as \emph{type I} and \emph{type II} errors. Indeed, it is easy to see that the two conditions in equation \eqref{eq:eo} (one for $y=1$ and one for $y=0$) are equivalent to requiring that the model has {\itshape the same false positive rate and false negative rate across groups identified via $A$}. False positives and false negatives are precisely type I and type II errors, respectively. Namely, individuals that are granted loans but are not able to repay, and individuals that are able to repay but are not granted loans.

\begin{figure}
    \centering
    \includegraphics[width=.4\textwidth]{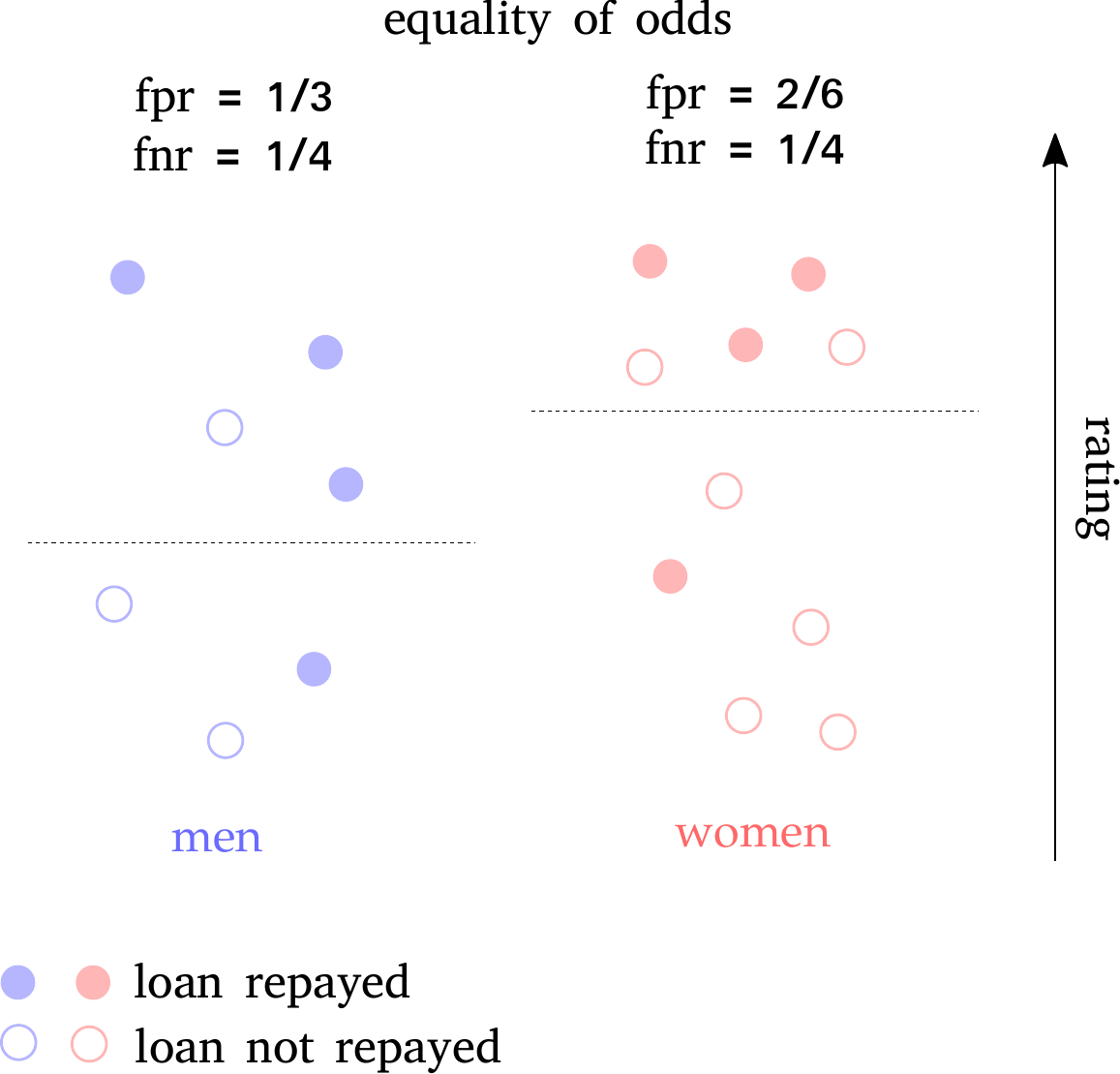}
    \caption{Example of Equality of Odds between men and women: false negative and false positive rates must be equal across groups.}
    \label{fig:EO}
\end{figure}

This is known as {\itshape Equality of Odds} \citep{hardt2016equality}, and is thus the requirement of having the same type I and type II error rates across relevant groups, as displayed in Figure \ref{fig:EO}.

There are two relaxed version of this criterion:
\begin{itemize}[-]
    \item {\itshape Predictive Equality}: equality of false positive rate across groups, 
    \begin{multline*}
        P(\hat{Y}=1\ \rvert\ A=a,Y=0) = \\P(\hat{Y}=1\ \rvert\ A=b,Y=0),\quad
        \forall a,b\in A,
    \end{multline*}
    \item {\itshape Equality of Opportunity}: equality of false negative rate across groups, 
    \begin{multline*}
        P(\hat{Y}=0\ \rvert\ A=a,Y=1) = \\P(\hat{Y}=0\ \rvert\ A=b, Y=1),\quad a,b\in \mathcal{A}.
    \end{multline*}
\end{itemize}

While demographic parity, and independence in general, focuses on equality in terms of acceptance rate (loan grating rate), Equality of Odds, and separation in general, focuses on equality in terms of error rate: the model is fair if it is as efficient in one group as it is in the other.

%As remarked above, in order to assess fairness in terms of model errors, one needs to trust the ground truth $Y$.  

The difference between Predictive Equality and Equality of Opportunity is the perspective from which equality is required: Predictive Equality takes the perspective of people that won't repay the loan, while Equality of Opportunity takes the one of people that will repay. Depending on the problem at hand, one may consider either of these two perspectives as more important. For example, Predictive Equality may be considered when we want to minimize the risk of innocent people from being erroneously arrested: in this case it may be reasonable to focus on the parity of among innocents. Both in the credit lending and job listing examples, on the other hand, it may be reasonable to focus on Equality of Opportunity, i.e. on the parity among people that are indeed deserving.

Here follows a non exhaustive set of situations in which separation criteria may be suitable:
\begin{itemize}[-]
    \item when your target variable $Y$ is an objective ground truth;
    \item when you are willing to make discrimination as long as they are justified by actual trustable data;
    \item when you do not want to actively enforce an ``ideal'' form of equality, and you want to be as equal as possible given the data.
\end{itemize}

We can summarize by saying that that separation, being a concept of parity given the ground truth outcome, is a notion that takes the point of view of people that are subject to the model decisions, rather than that of the decision maker. 
In the next subsection, instead, we shall take into account the other side of the coin, i.e. parity given the model decision.

\subsection{Sufficiency\label{sec:sufficiency}}

Sufficiency \citep{barocas-hardt-narayanan} takes the perspective of people that are given the same model decision, and requires parity among them irrespective of sensitive features.

While separation deals with error rates in terms of fraction of errors over the ground truth, e.g. the number of individuals whose loan request is denied among those who would have repaid, sufficiency takes into account the number of individuals who won't repay among those who are given the loan.

Mathematically speaking, this is the same distinction you have between recall (or true positive rate) and precision, i.e. $P(\hat{Y}=1\ \rvert\ Y=1)$ and $P(Y=1\ \rvert\ \hat{Y}=1)$, respectively.

A fairness criterion that focuses on this type of error rate is called {\itshape Predictive Parity}~\citep{chouldechova2017fair}, also referred to as \emph{outcome test}~\citep{verma2018fairness, mitchell2021algorithmic}:
\begin{multline}\label{eq:pp}
    P(Y=1\ \rvert\ A=a,\hat{Y}=1) =\\ 
    P(Y=1\ \rvert\ A=b,\hat{Y}=1),\quad \forall a,\ b\in \mathcal{A},
\end{multline}
i.e. the model should have the same precision across sensitive groups. 
If we require condition \eqref{eq:pp} to hold for the case $Y=0$ as well, then we get the following conditional independence statement: 
\[
    Y \indep A\ \rvert\ \hat{Y}, 
\]
which is referred to as {\itshape sufficiency} ~\citep{barocas-hardt-narayanan}.

Predictive Parity, and its more general form of sufficiency, focuses on error parity among people who are given the same decision. In this respect, Predictive Parity takes the perspective of the decision maker, since they group people with respect to the decisions rather than the true outcomes. 
Taking the credit lending example, the decision maker is indeed more in control of sufficiency rather than separation, since parity given decision is something directly accessible, while parity given truth is known only in retrospect. Moreover, as we have discussed above, the group of people who are given the loan ($\hat{Y} = 1$) is less prone to selection bias than the group of people who repay the loan ($Y = 1$): indeed we can only have the information of repayment for the $\hat{Y} = 1$ group, but we know nothing about all the others ($\hat{Y} = 0$).

As you may notice, going along a similar reasoning, one can define other group metrics, such as Equality of Accuracy across groups: $P(\hat{Y}=Y\ \rvert\ A=a) = P(\hat{Y}=Y\ \rvert\ A=b)$, for all $a,b\in \mathcal{A}$, i.e. focusing on over unconditional errors, and others~\citep{verma2018fairness}.

\subsection{Group fairness on scores\label{sec:score}}

In most cases, even in classification setting, the actual output of a model is not a binary value, but rather a \emph{score} $S\in\mathbb{R}$, estimating the probability, for each observation, to have the target equal to the favorable outcome (usually labelled 1).
Then, the final decision is made by the following $t$-threshold rule:
\begin{equation}
    \hat{Y} = \left\{\begin{aligned} 
        1\quad S \geq t,\\
        0\quad S < t.
    \end{aligned}\right.
\end{equation}

Most of the things we have outlined for group fairness metrics regarding $\hat{Y}$ can be formulated for the joint distribution $(A, Y, S)$ as well:
\begin{equation}
\label{eq:score}
\begin{aligned}
    &\text{independence} \quad S \indep A,\\
    &\text{separation} \quad S \indep A\ \rvert\ Y,\\
    &\text{sufficiency} \quad Y \indep A\ \rvert\ S.
\end{aligned}
\end{equation}
These formulations provide stronger constraints on the model with respect to the analogous with $\hat{Y}$, e.g.
\citet{jiang2020wasserstein, oneto2020fairness} call the condition $S \indep A$ \emph{strong demographic parity}.

For instance, all three criteria in their form \eqref{eq:dp}, \eqref{eq:eo}, \eqref{eq:pp}, i.e. with constraints on the joint distribution $(A, Y, \hat{Y})$ can effectively be satisfied by defining group dependent thresholds $t$ on black-box model outcomes $S$ (a technique that goes under the name of {\itshape postprocessing} \citep{barocas-hardt-narayanan}), while this is not the case for \eqref{eq:score}.

Instead of requiring conditions on the full distribution of $S$ as in \eqref{eq:score}, something analogous to the ``binary versions'' of group fairness criteria have been defined simply by requiring parity of the \emph{average score}~\citep{kleinberg2016inherent}.
\emph{Balance of the Negative Class}, defined as 
\begin{multline}
\label{eq:balance_neg}
    E(S\ \rvert\ Y = 0, A=a) = \\ 
    E(S\ \rvert\ Y = 0, A=b),\quad\forall a,b\in\mathcal{A}, 
\end{multline}
corresponds to Predictive Equality, while \emph{Balance of the Positive Class} (same as \eqref{eq:balance_neg} with $Y=1$) corresponds to Equality of Opportunity. Notice that these two last definitions fall into the one given in subsection~\ref{sec:separation} when $S = \hat{Y}$. Requiring both balances is of course equivalent to requiring EO.

\emph{AUC parity}, namely the equality of the area under the ROC for different groups identified by $A$, can be seen as the analogous of the equality of accuracy.

Finally, notice that, the score formulation of sufficiency is connected to the concept of {\itshape calibration}. Calibration holds when
\begin{equation}
    P(Y = 1\ \rvert\ S = s) = s,
\end{equation}
i.e. if the model assigns a score $s$ to 100 people then, on average, $100\times s$ of them will actually be positive.
Writing down the condition for sufficiency in score:
\begin{multline}
    P(Y = 1\ \rvert\ S = s, A = a) = \\
    P(Y = 1\ \rvert\ S = s, A = b),\quad\forall a,b \in \mathcal{A}, \forall s,
\end{multline}
it is clearly related to what can be called {\itshape Calibration within Groups}~\citep{kleinberg2016inherent}:
\begin{equation}
    P(Y = 1\ \rvert\ A = a, S = s) = s
\end{equation}
-- actually a consequence of it.
This condition is in general not so hard to achieve, and it is often satisfied ``for free'' by most models. See \citet{barocas-hardt-narayanan} for more details.

\subsection{Incompatibility statements\label{sec:incompatibilities}}

It is interesting to analyse the relationships among different criteria of fairness. 
In subsection~\ref{sec:groupvsindividual} we shall discuss in detail the connections between individual and group notions, while here we focus on differences among various  group criteria. We have already seen that each of them highlights one specific aspect of an overall idea of fairness, and we may wonder what happens if we require to satisfy multiple of them at once.
The short answer is that it is not possible except in trivial or degenerate scenarios, as stated by the following propositions drawn from the literature~\citep{chouldechova2017fair, kleinberg2016inherent, barocas-hardt-narayanan, raz2021group, simoiu2017problem}:

\begin{enumerate}
    \item \emph{if $Y$ is binary,  $Y \notindep S$ and $Y\notindep A$, then \textbf{separation} and \textbf{independence} are incompatible}.

    In other words, to achieve both separation and independence, the only possibility is that either the model is completely useless ($Y\indep S$), or the target is independent of the sensitive attribute  ($Y \indep A$), which implies an equal base rate for different sensitive groups.
    Namely, if there is an imbalance in groups identified by $A$, then you cannot have both EO and DP holding.
    
    \item Analogously: \emph{if $Y\notindep A$, then \textbf{sufficiency} and \textbf{independence} cannot hold simultaneously}.

    Thus, if there is an imbalance in base rates for groups identified by $A$, then you cannot impose both sufficiency and independence.
    
    \item Finally, \emph{if $Y \notindep A$ and the distribution $(A, S, Y)$ is strictly positive, then \textbf{separation} and \textbf{sufficiency} are incompatible}.

    Meaning that separation and sufficiency can both hold either when there is no imbalance in sensitive groups (i.e. the target is independent of sensitive attributes), or when the joint probability $(A, S, Y)$ is degenerate, i.e. --- for binary targets --- when there are some values of $A$ and $S$ for which only $Y=1$ (or $Y=0$) holds, in other terms when the score exactly resolves the uncertainty in the target (as an example, the perfect classifier $S=Y$ always trivially satisfies both sufficiency and separation).
\end{enumerate}
Notice that proposition 3 reduces to the following, more intuitive statement if the classifier is also binary (e.g. when $S=\hat{Y}$): \emph{if $Y \notindep A$, $Y$ and $\hat{Y}$ are binary, and there is at least one false positive prediction, then separation and sufficiency are incompatible.}
Moreover, it has be shown~\citep{kleinberg2016inherent} that \emph{Balance of Positive Class, Balance of Negative Class and Calibration within Groups can hold together only if either there is no imbalance in groups identified by $A$ or if each individual is given a perfect prediction (i.e. $S\in\{0, 1\}$ everywhere)}.

When dealing with incompatibility of fairness metrics, the literature often  focuses on the 2016 COMPAS\footnote{A recidivism prediction instrument developed by Northpoint Inc.} recidivism case~\citep{angwin2016machine}, now become a case study in the fairness literature. Indeed, the debate on this case is a perfect example to highlight the fact that there are \emph{different and non-compatible notions of fairness}, and that this may have concrete consequences on people. While we refer to the literature for a thorough discussion of the COMPAS case~\citep{washington2018argue, chouldechova2017fair}, we here just point out that in the debate there were two parties, one stating that the model predicting recidivism was \emph{fair} since it satisfied Predictive Parity by ethnicity, while the other claiming it was \emph{unfair} since it had different false positive and false negative rates for black and white individuals.  
\citet{chouldechova2017fair} showed that, if $Y \notindep A$, i.e. if the true recidivism rate is different for black and white people, then Predictive Parity and Equality of Odds cannot both hold, thus implying that a reflection on which of the two (in general of the many) notions is more important to be pursued in that specific case must be carefully considered.

Summarizing, apart from trivial or peculiar scenarios, the three families of group criteria above presented are not mutually compatible.

\subsection{Multiple sensitive features}

Even if a detailed discussion of the problem of multiple sensitive features is out of the scope of this manuscript, we shall nevertheless give a brief overview. 

Generally speaking, all the definitions and results we gave in previous sections are subject to the fact that the sensitive feature $A$ is represented by a single categorical variable. If, for a given problem, we identify more than one characteristic that we need to take into account as sensitive or protected -- say $(A_1, \ldots, A_l)$ -- we can easily assess fairness on each of them separately.  This approach, that \citet{yang2020fairness} call \emph{independent group fairness}, unfortunately is in general not enough: even if fairness is achieved (in whatever sense) separately on each sensitive variable $A_i$, it may happen that some subgroups given by the intersection of two or more $A_i$'s undergo unfair discrimination with respect to the general population. This is sometimes referred to as \emph{intersectional bias}~\citep{crenshaw1994mapping}, or, more specifically, \emph{fairness gerrymandering}~\citep{kearns2018preventing}.

To prevent bias from occurring in \emph{all the possible subgroups identified by all $A_i$'s} one can simply identify a new feature $A = (A_1, \ldots, A_l)$, whose values are the collection of values on all the sensitive attributes, and require fairness constraints on $A$. \citet{yang2020fairness} call this \emph{intersectional group fairness}. 

This last ``trick'' indeed solves the problem of intersectional bias, at least theoretically. Still, issues remain at a computational and practical level, whose two main reasons are: 
\begin{itemize}[-]
    \item the exponential increase of the number of subgroups when adding sensitive features,
    \item the fact that, with finite samples, many of the subgroups will be empty or with very few observations.
\end{itemize}
These two aspects imply that assessing (group) fairness with respect to multiple sensitive attributes may be unfeasible in most practical cases. Since the presence of many sensitive features is more of a norm than it is an exception, this actually represents a huge problem, that the literature on fairness in ML has barely begun to address~\citep{yang2020fairness, kearns2018preventing, kearns2019empirical,  buolamwini2018gender}.

\section{Group vs. individual fairness\label{sec:groupvsindividual}}

The most common issue with group fairness definitions is the following: since group fairness requires to satisfy conditions only on average among groups, it leaves room to bias discrimination \emph{inside the groups}. As we argued in the example of section~\ref{sec:group} referring to figure~\ref{fig:DP_1}, one way of reaching DP is to use a different rating threshold for men and women: this means that there will be a certain range of ratings for which men will receive the loan, and women won't. More formally, conditionally on rating there is no independence among $A$ and $\hat{Y}$. In general, to reach group fairness one may ``fine tune'' the interdependence of $A$ and $X$ to reach parity on average, but effectively producing differences in subgroups of $A$.

Notice that this is precisely what individual fairness is about. In the example above, a men and a woman that have the same rating may be treated differently, thus violating the individual notion of fairness.

As already discussed in section~\ref{sec:group}, of course this is only one possibility: one may as well reach DP by not using neither gender nor rating, and grant loan on the basis of other information, provided it is independent of $A$. In this case, there will be no group discrimination, but there won't be any subgroup discrimination as well.

However, we can say that, if we want to reach DP by using as much information of $Y$ as possible  contained in $(X, A)$, i.e. minimizing the risk $E\mathcal{L}(f(X, A), Y)$, then it is unavoidable to have some form of disparate treatment among people in different groups with respect to $A$ whenever $X \notindep A$.
This has been thoroughly discussed by \citet{dwork2012fairness}, where they call \emph{fair affirmative action} the process of requiring DP while trying to keep as low as possible the amount of disparate treatment between people having similar $X$.

\begin{figure}
    \centering
    \includegraphics[width = .4\textwidth]{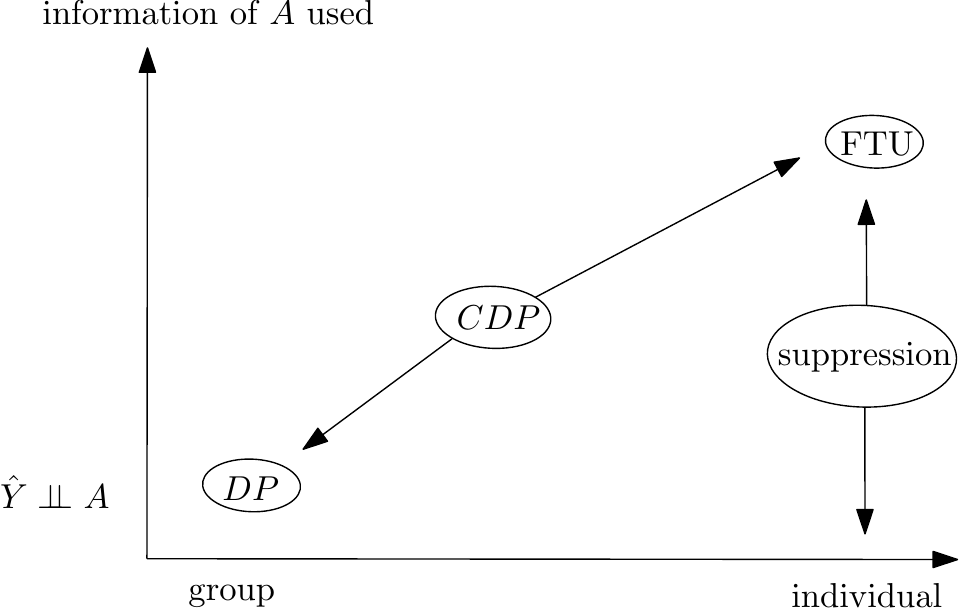}
    \caption{Landscape of observational fairness criteria with respect to the group-vs-individual dimension and the amount of information of $A$ used (via $X$).}
    \label{fig:observational_legitimate}
\end{figure}

To clarify the general picture, we can put the different notions of (observational) fairness in a plane with two \emph{qualitative} dimensions (see figure~\ref{fig:observational_legitimate}): 1. to what extent a model is fair at the individual level, 2. how much information of $A$ is retained in making decisions. The first dimension represents to what extent two individuals with similar overall features $X$ are given similar decisions: the maximum value is reached by models blind on $A$ (FTU). These are the models that are also using all information in $X$, irrespective of the interdependence with $A$, thus FTU-compliant models will use all information contained in $\widetilde{X}$ apart from the information that is contained in $A$ only. The minimum value in this dimension is reached by models that satisfy DP. Models using suppression methods, being blind to both $A$ and other features with high correlation with $A$,  are individually fair in the sense of preventing disparate treatment. In so doing, they can exploit more or less information of $A$ with respect to general DP-compliant models depending on how many correlated variables are discarded. However, the price to pay for discarding variables is in terms of errors in approximating $Y$, which is not highlighted in this plot. Notice that, of course, full suppression -- i.e. removing all variables dependent on $A$ -- trivially satisfies the condition $\hat{Y}\indep A$, i.e. it is DP-compliant as well. In other terms, one can have a DP-compliant model that is individual as well. In figure~\ref{fig:observational_legitimate}, we label with DP a general model that tries to maximize performance while satisfying a DP constraint, without any further consideration.

Models satisfying CDP are somewhat in-between, of course depending on the specific variables considered for conditioning. They guarantee less disparate treatment than unconditional DP, and they use more information of $A$ by controlling for other variables possibly dependent on $A$.

Notice that approaches such as fair representation (see section~\ref{sec:individual}), where you try to remove all information of $A$ from $X$ to get new variables $Z$ which are as close as possible to being independent of $A$, produce decision systems $\hat{Y} = f(Z)$ that are not, in general, individually fair. This is due to the simple fact that, precisely to remove the interdependence of $A$ and $X$ while keeping as much information of $X$ as possible, two individuals with same $X$ and different $A$ will be mapped in two distinct points on $\mathcal{Z}$, thus having, in general, different outcomes. Referring again to the credit lending example, suppose that we have $R = g(A) + U$, with $g$ a complicated function encoding the interdependence of rating and gender, and $U$ some other factor independent of $A$ representing other information in $R$ ``orthogonal'' to $A$. In this setting, the variable $Z$ that we are looking for is precisely $U$. Notice that $U$ is indeed independent of $A$, thus any decision system $\hat{Y} = f(U)$ satisfies DP, but given two individuals with $R = r$ nothing prevents them from having different $U$. 
In other terms, you \emph{need} to have some amount of disparate treatment to guarantee DP \emph{and} employ as much information as possible to estimate $Y$.

\begin{figure}
    \centering
    \includegraphics[width = .4\textwidth]{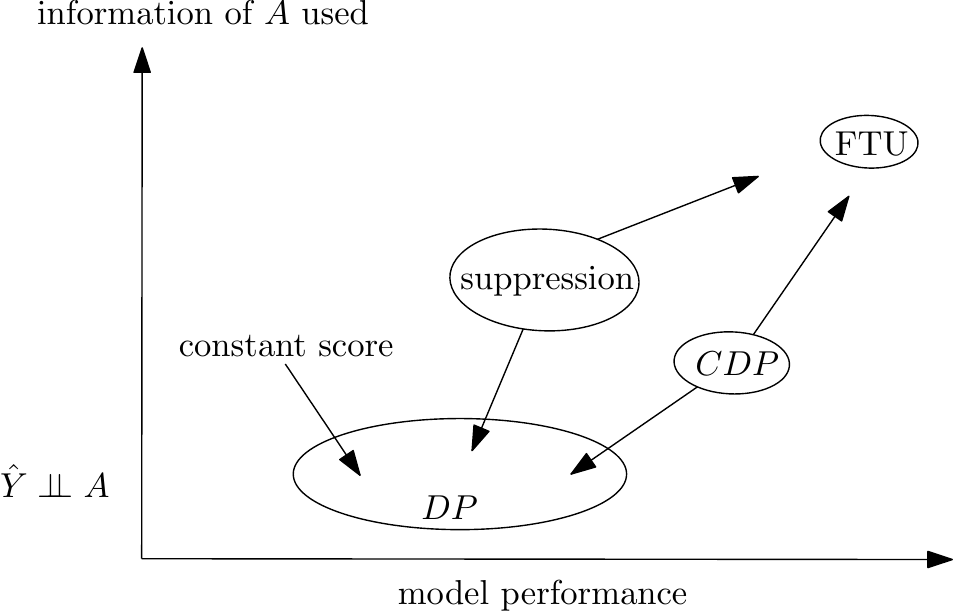}
    \caption{Landscape of observational fairness criteria with respect to the model performance dimension and the amount of information of $A$ used (via $X$).}
    \label{fig:performance}
\end{figure}

Figure~\ref{fig:performance} shows a \emph{qualitative} representation of observational metrics with respect to the amount of information of $A$ (through $X$) that is used by the model, and the predictive performance. Notice that DP can be reached in many ways: e.g. a constant score model, namely a model accepting with the same chance all the individuals irrespective of any feature, is DP-compliant (incidentally, it is also individual), a model in which all  the variables dependent on $A$ have been removed (a full suppression), or a model where DP is reached while trying to maximize performance (e.g. through fair representations). All these ways differ in terms of the overall performance of the DP-compliant model. 

FTU-compliant models, on the other hand, by employing all information in $X$ will be, in general, more efficient in terms of model performance.\footnote{Of course it is understood that the model $f$ in $\hat{Y} = f(X)$ is trained to maximize performance. }

Incidentally, notice that this discussion is to be taken at a qualitative level, one can come up with scenarios in which, e.g., models satisfying DP have higher performances than models FTU-compliant (think, e.g., of a situation in which $Y \indep A$ and $X \notindep A$).

The (apparent) conflict that hold in principle between individual and group families of fairness notions is debated in the literature~\citep{friedler2016possibility, hertweck2021moral, binns2020apparent}. Our analysis provides arguments to the claim that individual and group fairness are not in conflict \emph{in general}~\citep{binns2020apparent}, since they lie on the same line whose extreme values are separation (i.e. unconditional independence) and FTU (i.e. individuals are given the same decisions on the basis of non-sensitive features only), with conditional metrics ranging between them. 
Thus, the crucial aspect is assuming and deciding \emph{at the ethical and legal level} what are the variables that are ``allowed'' in specific scenarios: given those, we place ourselves on a point on this line reachable both by choosing an appropriate distance function on feature space (i.e. employing an individual concept) and by requiring parity among groups conditioned on those variables (a group notion).  

Indeed, when referring to the conflict between individual and group families one is committing a slight ``abuse of notation'', since the actual clash is rather on the \emph{assumptions} regarding what is to be considered fair in a specific situation, than on metrics \emph{per se}. Namely, one usually consider individual (group) concepts as the ones in which more (less) interdependence of $X$ and $A$ is allowed to be reflected in the final decisions. In the credit lending example, if income is correlated with gender, the issue weather it is fair to allow for a certain gender discrimination as long as justified by income can be seen as an instance of the conflict between individual or group notions, but it is rather a conflict about the underlying ethical and legal assumptions~\citep{binns2020apparent}.

\ref{sec:experiments} contains the description of a set of experimental results in support of the analysis outlined in this section.

\section{Causality-based criteria\label{sec:causality}}

Another important distinction of fairness criteria is the one between {\itshape observational} and {\itshape causality-based} criteria.

As we have seen, observational criteria rely only on observed realizations of the distribution of data and predictions. In fact, they focus on enforcing equal metrics (acceptance rate, error rate, etc\ldots) for different groups of people. In this respect, they don't make further assumptions on the mechanism generating the data and suggest to assess fairness through statistical computation on observed data. 

Causality-based criteria, on the other hand, try to employ domain and expert knowledge in order to come up with a casual structure of the problem, through which it becomes possible to answer questions like ``what would have been the decision if that individual had a different gender?''. 

While counterfactual questions like this seem in general closer to what one may intuitively think of as ``fairness assessment'', the observational framework is on the one hand easier to assess and constrain on, and on the other more robust, since counterfactual criteria are subject to strong assumptions about the data and the underlying mechanism generating them, some of which are not even falsifiable.

As we argued above (section~\ref{sec:individual}), answering to counterfactual questions is {\itshape very different} from taking the feature vector of, e.g., a male individual and just flip the gender label and see the consequences in the outcome. The difference lies precisely in the causal chain of ``events'' that this flip would trigger. If there are some features related, e.g., to the length of the hair, or the height, then it is pretty obvious that the flip of gender should come together with a change in these two variables as well. And this may be the case for other, less obvious but more important, variables.

This also suggests why counterfactual statements involve {\itshape causality relationships} among the variables. In general, to answer counterfactual questions, one needs to know the causal links underlying the problem. This requires a certain number of assumptions, usually driven by domain knowledge. 

However, as major drawback, once given the casual structure there are many counterfactual models compatible with that structure (actually infinite), and the choice of one of them is not falsifiable in any way. 

Indeed, causality-based criteria can be formulated at least at two different levels, with increasing strength of assumptions: at the level of \emph{interventions} and at the level of \emph{counterfactuals}.

\subsection{Causal models}

Before introducing fairness definitions based on the underlying causal structure of a problem, we briefly introduce the necessary theoretical framework of \emph{causal graphs} and \emph{Structural Causal Models} (SCMs) or Structural Equation Models~\citep{pearl2018book, pearl2016causal, peters2014causal,barocas-hardt-narayanan,kusner2017counterfactual}.

\begin{figure*}
\begin{center}
\subfigure{
\begin{tikzpicture}
    % nodes
    %% endogenous
    \node[state] (x) at (-1.5, 1) {$X$};
    \node[state] (a) at (-1.5, -1) {$A$};
    \node[state] (y) at (1.5, 0) {$Y$};
    %% exogenous
    \node[state, fill=gray!30, dashed] (Ux) at (-3, 1) {$U_X$};
    \node[state, fill=gray!30, dashed] (Ua) at (-3, -1) {$U_A$};
    \node[state, fill=gray!30, dashed] (Uy) at (2, 1) {$U_Y$};
    % edges
    \path (x) edge (a);
    \path (x) edge (y);
    \path (a) edge (y);
    \path[dashed] (Ux) edge (x);
    \path[dashed] (Ua) edge (a);
    \path[dashed] (Uy) edge (y);
\end{tikzpicture}
}\qquad
\subfigure{
\begin{tikzpicture}
    % nodes
    %% endogenous
    \node[state] (x) at (-1.5, 1) {$X$};
    \node[state] (a) at (-1.5, -1) {$A=a$};
    \node[state] (y) at (1.5, 0) {$Y$};
    %% exogenous
    \node[state, fill=gray!30, dashed] (Ux) at (-3, 1) {$U_X$};
    \node[state, fill=gray!30, dashed] (Uy) at (2, 1) {$U_Y$};
    % edges
    \path (x) edge (y);
    \path (a) edge (y);
    \path[dashed] (Ux) edge (x);
    \path[dashed] (Uy) edge (y);
\end{tikzpicture}
}
\end{center}
\caption{Example of causal graph with 3 endogenous variables (left). Intervention on $A$ is expressed via a different graph (right) where all incoming edges in $A$ are removed and the variable $A$ is set to the value $a$.}
\label{fig:causal_graph}
\end{figure*}
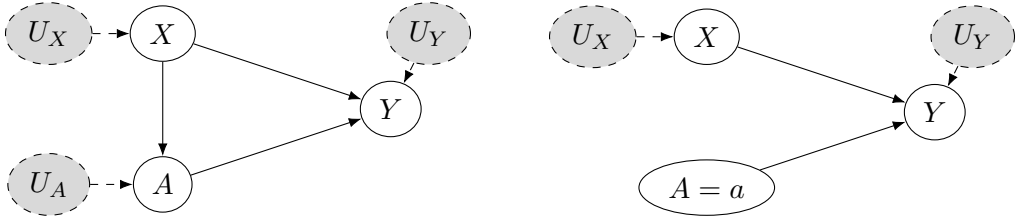

We model the underlying causal relationships among features by means of a Directed Acyclic Graph (DAG) $G = (V, E)$, with $V$ set of vertices (or nodes) and $E$ set of directed edges (or links) --- see figure~\ref{fig:causal_graph}. Nodes of the graph $G$ represent the variables $\w{X}$ used as predictors in the model. Moreover, we denote with $U$ exogenous or \emph{latent} variables, representing factors not accounted for by the features $\w{X}$. In causal graph theory, edges in the graph represent not only conditional dependence relations, but are interpreted as the causal impact that the source variable has on the target variable. 
To work at the level of \emph{interventions} this is all that is needed.
Modeling causal knowledge is complex and challenging since it requires an actual understanding of the relations, beyond statistical evidence. Different causal discovery algorithms have been proposed to identify causal relationships from observational data through automatic methods~\citep{glymour2019review}.

To work at the level of \emph{counterfactuals} one needs much more than the sole graph structure: a Structural Causal Model is a a triplet $(\w{X}, U, F) $ where $F$ are a set of assignments of the form
\begin{equation}
    \w{X}_i = f_i(\pa(\w{X}_i), U_i),\quad i = 1, \ldots, d;
\label{eq:SE}
\end{equation}
where the functions $f_i$ represent the precise way in which the parents of each node variable $\w{X}_i$  ($\pa(\w{X}_i)$) --- namely the variables having a direct causal impact over $X_i$ in the assumed (or inferred) graph $G$ --- together with the latent variable $U_i$, influence the value of $X_i$. These relations are called Structural Equations (SE) and, besides describing which variables causally impacts which (that is already encoded in the graph $G$), they also determine \emph{how} these relations work. 
One typical simplifying assumption on SCM is to work with Additive Noise Models~\citep{peters2014causal}:
\[
\w{X}_i = g_i(\pa(\w{X}_i)) + U_i,\quad i=1,\ldots d.
\]
In general, assumptions like this are needed in order to be able to perform computations using SCM starting from observational data.
A detailed presentation of the assumptions underlying the use of graphical models and SCMs (e.g. modularity, markovianity, causal sufficiency, etc\ldots) is beyond the scope of this survey~\citep{peters2014causal, guo2020survey, dawid2010beware, pearl2016causal}.  

As mentioned above, one of the limits of counterfactual models is that, since they make ``predictions'' regarding something that has not happened, the choice of a specific counterfactual model is in general not falsifiable with observable data~\citep{peters2014causal, dawid2000causal}.

\subsection{Fairness metrics in causality-based settings}

Fairness at the level of interventions can be formally expressed as follows~\citep{kilbertus2017avoiding}:
\begin{multline}
\label{eq:intervention}
    P(\hat{Y} = 1\ \rvert\ do(A = a), X=x) = \\P(\hat{Y} = 1\ \rvert\ do(A = b), X=x),\quad
    \forall a, b \in \mathcal{A}, x\in\mathcal{X}.
\end{multline}
In words, if you take a random individual, force it to be, e.g., female ($do(A=a)$) and she happens to have $X=x$, you want to give him the same chance of acceptance as for a random individual forced to be male ($do(A = b)$) that also happens to have $X=x$\footnote{It may be unfeasible to compute quantities in \eqref{eq:intervention} on the basis of observational data and the graph, without assuming an SCM (or without physically performing the experiment). Indeed it may be that the events whose probability needs to be computed in order to get the post-intervention distribution have no observations in the dataset at hand~\citep{pearl2016causal, pearl2009causality}}. We refer to it as \emph{Intervention Fairness}. 

The same requirement can be set at the counterfactual level, and is known as \emph{Counterfactual Fairness (CFF)}~\citep{kusner2017counterfactual}:
\begin{multline}
\label{eq:cff}
    P(\hat{Y}_{A\leftarrow a}=1\ \rvert\ A=a,X=x) = \\
        P(\hat{Y}_{A\leftarrow b}=1\ \rvert\ A=a,X=x),\quad
    \forall a,b\in \mathcal{A},
\end{multline}
which, in words, reads: if you take a random individual with $A=a$ and $X=x$ and the same individual if she had $A = b$, you want to give them the same chance of being accepted.  

The difference between the two levels is subtle but important: roughly speaking, when talking about interventions one is considering the average value over exogenous factors $U$ that, after the intervention, are compatible with the conditioning, while counterfactuals consider only the values of $U$ that are compatible with the factual observation (namely, the distribution P$(U\ \rvert\ A=a, X=x)$) to begin with, and then perform the intervention and consider the consequences. In other words, counterfactuals consider only events that take into account actual observed value of $A$ (and $X$ as well). As a further clarification~\citep{kusner2017counterfactual}, suppose the following structural equation: $X=A+U$. Then equation \eqref{eq:intervention} compares two individuals with $U=x-a$ and $U'=x-b$, i.e. two different individuals that, with the interventions $do(A=a)$ and $do(A=b)$ both happen to have the observed value $X=x$. 
Equation~\eqref{eq:cff} instead take the individual with $U=x-a$ and then perform the interventions $do(A=a)$ (which has no consequences) and $do(A=b)$ which implies on $X = b + (x-a) \neq x$.

Equations~\eqref{eq:intervention}, \eqref{eq:cff} make use of Pearl's $do$-calculus~\citep{pearl2009causality, pearl2016causal, pearl2018book}, where $do(A = a)$ --- the \emph{intervention} --- consists in modifying the causal structure of the problem by exogenously setting $A = a$, thus removing any causal paths impacting on $A$ -- the theoretical equivalent of randomized experiments (see figure~\ref{fig:causal_graph} (right)). Effectively, post-intervention variables have a different distribution $P_{do(A=a)}(X, A, Y)$, and samples (or observations) from this distribution can be drawn only if the intervention $do(A=a)$ can be physically carried out (e.g. in randomized experiments). In many real life situations this is not feasible, and $do$-calculus provide a mathematical machinery that allows, in some specific cases, to compute quantities related to the (non observable) post-intervention distribution $P_{do(A=a)}(X, A, Y)$ employing only quantities related to the (observable) distribution $P(X, A, Y)$, given the graph --- a concept called \emph{identifiability}~\citep{pearl2009causality, pearl2018book, pearl2016causal, peters2014causal, guo2020survey}.

The notation $\hat{Y}_{A \leftarrow b}$, on the other hand, stands for the three steps of counterfactual calculus, \emph{abduction, action, prediction}\cite{pearl2016causal, pearl2018book, pearl2009causality}: 
\begin{itemize}[-]
    \item \emph{abduction} is where you account for observed values and compute the distribution of $U\ \rvert\ \{A=a, X=x\}$; 
    \item \emph{action} corresponds to implementing the intervention $do(A = b)$;
    \item \emph{prediction} consists in using the new causal structure and the exogenous conditional distribution P$(U\ \rvert\ A=a, X=x)$ to compute the posterior of $\hat{Y}$.
\end{itemize} 
 We refer to \citet{pearl2018book, pearl2009causality, pearl2016causal} for a general review on causal inference and $do$-calculus, and again to \citet{barocas-hardt-narayanan} for a thorough discussion of causality in the context of fairness.

As in the observational setting, causality-based criteria have a group and an individual version: equations~\eqref{eq:intervention}, \eqref{eq:cff} correspond to the individual form, but nothing prevents to take the version unconditional on $X$, i.e. holding on average on all the individuals, or even to condition on only some subset of $X$, as in CDP.

\subsection{Group vs. individual fairness in causality-based setting}
\label{sec:individual_causal}

We call \emph{Expectation Intervention Fairness} the condition:
\begin{multline}
    P(\hat{Y}=1\ \rvert\ do(A=a)) =\\
    P(\hat{Y}=1\ \rvert\ do(A=b))\quad \forall a,b \in \mathcal{A},
\end{multline}
i.e. requiring that, on average, an individual taken at random from the whole population and forced to be woman ($do(A=a)$) should have the same chance of being accepted as a random individual forced to be man.
Analogously, we can define \emph{Expectation Counterfactual Fairness (ECFF)} as:
\begin{multline}
    P(\hat{Y}_{A\leftarrow a}=1\ \rvert\ A=a) =\\
    P(\hat{Y}_{A\leftarrow b}=1\ \rvert\ A=a)\quad \forall a,b \in \mathcal{A},
\end{multline}
which states that, on average, the acceptance rate for a random woman ($A=a$) should be the same as the one given to a random woman forced to be a man.
Similar definitions can be given conditioning on partial information $R$.

Summarizing, we can visualize CFF as the following process: the conditioning on some $(A=a, R=r)$ represents the group of people that we take into account (a single individual for $R=X$), and any consideration we do on them must hold on average over that group; given that group, we force a flipping of the sensitive attribute from $A=a$ to $A=b$ (and here we are going in a new, non-observable, distribution), and this will trigger a cascade of causal consequences on the other features $X$ (namely, on the descendants of $A$ in the causal graph). Then, we compare the model outcomes averaged on the observed group and on the counterfactual group. 

Intervention Fairness is slightly different: you take again all the individuals compatible with the conditioning -- which this time is $R=r$ without fixing the sensitive $A$ -- then force them first to have $A=a$, and then to have $A=b$ (with all the causal consequences of these interventions), and require them to have, on average, the same acceptance rate.

This may resonate with the notion of FTU: when flipping $A$ for an observation you want the outcome of the model not to change. Indeed \emph{Fairness Through Unawareness is a causality-based notion of fairness}, at least in its formulation of not explicitly using $A$. The very concept of flipping $A$ is nothing but an intervention. Notice, incidentally, that the flipping of $A$ \emph{without any impact on other variables} corresponds to assuming a causal graph where $A$ has no descendant, or, better, corresponds to assuming that all the changes caused by the flipping of $A$ on other variables are legitimate, i.e. considered fair. Moreover, the fact that FTU is difficult to be measured without having access to the model, is due to its nature of being a non-observational notion, but it requires fictitious data to be assessed -- a dataset with $A$ flipped, i.e. a \emph{dataset not sampled from the real distribution of $(A, X, \hat{Y})$}. This is similar to the way in which CFF can be assessed: compare the predictions over a dataset with the predictions on the same dataset but with $A$ flipped \emph{and} with all the changes caused by this flipping\footnote{Knowledge of the underlying Structural Causal Model is needed.}.
This, in turn, reveals the subtle difference among CDP in the form $\hat{Y}\indep A\ \rvert\ X$ and FTU: CDP is an observational notion, i.e. can be measured, in principle, by having access to (realizations from) the distribution of $(\hat{Y}, A, X)$, while FTU does not. We say ``in principle'' since it's very likely that in real-world datasets there will be very few observations corresponding to $X=x$ (typically one), thus resulting in a poor estimation of the distribution of $\hat{Y}\ \rvert\ X$.
%This, again, resonates to the considerations we did about the metric in equation~\eqref{eq:consistency} and its potential flaw of not capturing FTU, since in the neighborhood of a point with $A=a$ there may not even be a point with $A=b$ to compare to. This is why we need to build a fake dataset to assess FTU, because, ultimately, we want to compare the real-world distribution with a new (fake) distribution.

Causality-based notions are richer than the observational ones, and permit a further possibility: to select which causal paths from $A$ to $\hat{Y}$ are considered legitimate and which, instead, we want to forbid. In the job listing example, we may consider that the type of degree of the applicants is fundamental for the job position, and thus crucial to making decisions. But it could well be that the type of degree is correlated, and even ``caused'' by the gender of the applicants: women and men may have different attitudes towards choosing the preferred degree. In this situation, if we require CFF, we would compare a man with some degree to his ``parallel self'' in the counterfactual world where he is a woman. In that world, however, it's very likely that her degree is going to be different. Thus, requiring CFF means to somehow prevent the decision maker to employ the degree type for the assessment. 
Taking into account situations like this one is not very difficult. Supplementary Material of \citet{kusner2017counterfactual} and \citet{chiappa2019path} introduce the following definition, known as \emph{path-specific Counterfactual Fairness (pCFF)}:
\begin{multline}
\label{eq:pcff}
    P(\hat{Y}_{A\leftarrow a, X_F\leftarrow x_1}=1\ \rvert\ A=a,X_F=x_1, X_F^c=x_2) = \\
        P(\hat{Y}_{A\leftarrow b, X_F\leftarrow x_1}=1\ \rvert\ A=a,X_F=x_1, X_F^c=x_2),\\
    \forall a,b\in \mathcal{A}, \forall x_1, x_2,
\end{multline}
where $X_F$ correspond to the variables descendants of $A$ that we consider fair mediators to make decisions (e.g. the degree type), and $X_F^c$ is its complement with respect to $X$, namely $X = (X_F, X_F^c)$. In words, take a man ($A=a$) with a specific degree $x_1$ and other features $x_2$, and force him to be a woman, letting all the causal consequences of this flipping to happen \emph{with the exception of the degree, kept fixed at $x_1$}, then compare their outcomes.

Incidentally, notice that if we allow $X_F$ to contain all the descendants of $A$, then we end up with a notion that we may call \emph{direct Counterfactual Fairness (dCFF)}, namely the only path that we are concerned of is the direct causal path from gender $A$ to the decision $\hat{Y}$. In this case we don't allow any causal consequences of the gender flipping to happen when assessing fairness. This is, again, strictly connected to FTU.

\begin{figure}
    \centering
    \includegraphics[width = .4\textwidth]{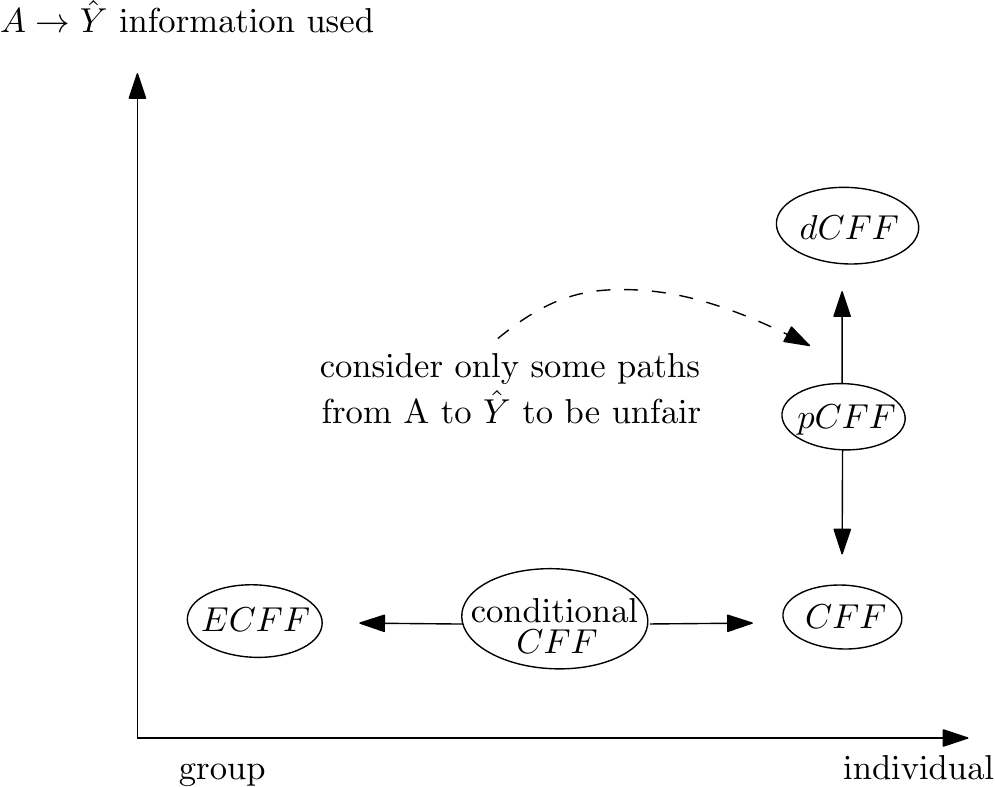}
    \caption{Landscape of causality-based fairness criteria.}
    \label{fig:causal}
\end{figure}

Figure~\ref{fig:causal} is a schematic and qualitative visualization of many of the points discussed above: the dimension of group vs. individual is ``controlled'' by how much information we condition on, while the dimension of the causal impact of $A$ on the decision is controlled  by the fraction of paths causally connecting $A$ to $\hat{Y}$ that are considered fair. Figure~\ref{fig:causal} is meant to be the causal analogue of figure~\ref{fig:observational_legitimate}. Of course there are many possible cases that, for simplicity, we have omitted from figure~\ref{fig:causal}, namely all the Intervention notions, and the path-specific notions valid on average in broader groups.\footnote{For instance,  one could think of a measure taking into account only the $A$ flip, without causal consequences, and valid on average over $X$ as well (i.e. a group notion), namely
\begin{multline}
    P(\hat{Y}_{A\leftarrow a, X\leftarrow X}=1\ \rvert\ A=a) = \\
        P(\hat{Y}_{A\leftarrow b, X\leftarrow X}=1\ \rvert\ A=a),\quad
    \forall a,b\in \mathcal{A}.
\end{multline}} 

If we focus on CFF, it can be reached by training a model on the space $(\Sigma, U_A, Y)$ where $\Sigma$ are the \emph{non-descendants} of $A$, i.e. variables not caused by $A$, \emph{directly or indirectly}; $U_A$ denotes the information inside descendants of $A$ that is not attributable to $A$. 
A decision system $\hat{Y} = f(\Sigma, U)$ is counterfactually fair~\eqref{eq:cff} by design.
This closely reflects what we discussed in section~\ref{sec:individual} about fair representations: train a model on a new space ``cleaned'' from all $A$ information. Indeed, in that case we searched for statistical independence of $\hat{Y}$ on $A$ while here we look for ``causal independence'' of $\hat{Y}$ on $A$. In this specific sense, counterfactual fairness can be reached via a form of \emph{preprocessing} of the dataset, and is very similar, in spirit, to the concept of fair representation.

We may then wonder why the fair representation approach, reaching DP, is a notion of group rather than individual fairness, while CFF is considered an individual notion: the point is simply that \emph{we employed two different concepts of individual fairness in the observational and in the counterfactual setting}. Namely, in the observational setting we expressed disparate treatment as the event in which two individuals with same $X$ but different $A$ are given different outcomes. In general, fair representation is not guaranteed to prevent such scenario. In the counterfactual context, instead, we consider disparate treatment when an individual and her ``counterfactual self'' with $A$ flipped are given different outcomes. CFF is designed precisely to prevent this. 
Notice that this concept, translated in the observational setting, is analogous to requiring a similar treatment for two individuals similar in $\mathcal{Z}$, not in $\mathcal{X}$. Indeed, if you take an individual with some $(A = a, X=x)$ and flip her gender \emph{while taking into account the interdependence of $X$ and $A$}, she will be transformed in the same point $Z$. 

This is an example of the fact that the concept of individual fairness is strongly dependent on the concept of similarity that one decides to consider.

Summarizing, CFF and fair representation are very similar in the way in which they deal with disparities, namely they both try to remove all (causal) information of $A$ from the feature space, and then use the ``cleaned'' space to make decisions. In this respect, CFF can be seen as both a ``causal analogue'' of a DP-compliant methods, in the sense that all the way in which $A$ may impact the decision are forbidden, and an individually fair notion in the sense that it imposes condition on individual basis.

\section{Conclusions}

The notion of fairness in decision making has many nuances, that have been reflected in the high number of proposed mathematical and statistical definitions. Notice, moreover, that this aspect is not limited to ML or artificial intelligence: the problem of how to define and assess bias discrimination in decision making processes is present largely independently of ``who'' is making the decision. The growing attention to this issue in the domain of automated data-driven decisions can be imputed to the fact that these processes can potentially amplify biases \emph{at scale}, and could possibly do that without \emph{human oversight}. It is well known that AI systems usually compute their outcomes leveraging very complicated relationships among input features, making not only the users but often also the developer blind with respect to the reasons underlying those outcomes. This issue, that has triggered in the recent years the flourishing stream of research of Explainable AI~\citep{guidotti2018survey, burkart2021survey}, is intertwined to that of bias detection and fairness in ML, since the problem of discrimination becomes even subtler when carried out by a ``black-box'' algorithm whose rationale is obscure or ambiguous.

Even if a lot of work has been done in this respect, still there is confusion on the interplay among different fairness notions and metrics to assess them. In this paper, we tried to highlight some important aspects about the relationships between fairness metrics, in particular with respect to the clash ``individual vs. group'' and ``observational vs. causality-based'' and we have argued that we are facing rather a landscape of intertwined possibilities than clear-cut distinctions.

Some have expressed critics and doubts about the possibility of capturing the complexity of notions such as equity and fairness with quantitative methods~\citep{green2018myth}. Even if these doubts are reasonable, we believe that quantitative research in the domain of fairness notions can provide an important contribution to the more general issue of bias discrimination, at least in terms of understanding and comprehension of the problems and possibly to highlight shortcomings and subtleties -- let alone to foster a boost in the attention of the scientific community on it. Thus, we strongly welcome more research both on the quantitative aspects, dealing with assessment metrics and algorithmic bias mitigation, and on the societal and legal aspects, and hopefully in the fruitful interplay between them.

\section*{Author contributions}
A.C., R.C., G.G. and D.R. conceived the main ideas of the paper; R.C. and D.R. worked out the details; D.R. wrote the manuscript with the help of R.C.; A.C. and G.G. checked and reviewed the manuscript; I.G.P. and A.C.C. reviewed the manuscript.

%% Bibliography %%
\bibliography{references.bib}

%% appendix %%
\appendix
\section{Experiments\label{sec:experiments}}

In order to support the qualitative insights presented in section~\ref{sec:groupvsindividual}, we perform experiments on three different dataset: a widely used public dataset of the financial domain (known as the Adult dataset)~\citep{Dua:2019}, and two synthetic datasets designed to capture different aspects of the manifold possibilities that may arise with minimal structure. For each dataset we measure several metrics on four different algorithmic approaches.

For the synthetic datasets we employ the following generative model, consisting of: two non-sensitive continuous features $X_1$ and $X_2$; a binary non-sensitive feature $X_3$; a binary sensitive attribute $A$: and two different target variables, $Y_h$ and $Y_l$, corresponding to two distinct experiments with a high and low correlation with $A$, respectively:
\begin{subequations}
\begin{align}
    &A \sim \text{Ber}(1/2);\\
    &X_1 = A/2 + U_1,\quad U_1 \sim \text{Norm}(0, 1/2);\label{eq:X1} \\ 
    &X_2 = U_2,\quad U_2 \sim \text{Norm}(0, 1/2);\\
    &X_3 = \bm{1}_{(A + U_3) \geq 1} ,\quad U_3 \sim \text{Ber}(1/2);\label{eq:X3} \\ 
    & \\
    &\begin{aligned}
        &U_{\zeta_h} \sim \text{Norm}(0, 1/2),\\
        &\zeta_h = X_1 + 2X_2 + X_3/2 + 4A + U_{\zeta_h},\\
        &Y_h = \bm{1}_{\zeta_h > \mathbb{E}(\zeta_h)};\\\label{eq:zeta_h}
    \end{aligned}\\ 
    &\begin{aligned}
        &U_{\zeta_l} \sim \text{Norm}(0, 1/2)\\
        &\zeta_l = X_1 + 2X_2 + X_3/2 + U_{\zeta_l},\\
        &Y_l = \bm{1}_{\zeta_l > \mathbb{E}(\zeta_l)};\label{eq:zeta_l}
    \end{aligned}
\end{align}
\label{eq:synth}
\end{subequations}
where $\bm{1}_{A}$ denotes the indicator function of the set $A$.
The interdependence between sensitive and non-sensitive features is governed by equations~\eqref{eq:X1}, \eqref{eq:X3}; the target variable $Y_l$ (low correlation with $A$) depends on $A$ via the $X_3/2$ term, while $Y_l$ (high correlation with $A$) has an additional $4A$ term precisely to increase the dependence of $Y$ on $A$. The rationale of model \eqref{eq:synth} is the following: $X_1$ and $X_2$ represent 2 continuous valued variables dependent and independent of $A$ respectively, introduced to have a minimal set of variables with non-trivial dependence on $A$; $X_3$ is the variable that we shall employ to compute CDP --- i.e. with respect to whose groups we want to impose acceptance rate parity --- thus we need a categorical variable with non-trivial dependence on $A$. Target variables $Y_h$ and $Y_l$ are simple linear combinations of features, in particular $Y_l$ does not depend explicitly on $A$.

We generate 15,000 samples from the model~\eqref{eq:synth} two times, one with target variables $Y_h$ and one with target $Y_l$, and label them synthetic $\# 1$ and synthetic $\# 2$, respectively. 
The Adult dataset~\citep{Dua:2019} is composed by 32,561 observations of census data, with 14~attributes, and the binary target denotes whether income exceeds \$50K/yr. We employ gender as sensitive attribute.

% experiments results
\begin{table*}[ht!]
\caption{\textbf{Results of experiments.} Normalized Mutual Information $U$, fairness (DP-ratio, Flip) and performance (ROC~AUC) metrics for different algorithmic approaches described in section~\ref{sec:experiments} on 3 datasets: the Adult public dataset and 2 synthetic datasets generated by samples of model~\eqref{eq:synth}. All metrics are expressed in percentage values.}
\label{tab:experiment}
\centering
\ra{1.4}
\resizebox{.99\textwidth}{!}{
\begin{tabular}{@{}lcccccccccccccccccc@{}}
  \toprule
  
  & \phantom{abc} & \multicolumn{5}{c}{\textbf{Adult}} & \phantom{abc} &  \multicolumn{5}{c}{\textbf{synthetic \#1}} & \phantom{abc} & \multicolumn{5}{c}{\textbf{synthetic \#2}} \\
  
  && \multicolumn{5}{c}{$U(Y, A) = 4.5$} && \multicolumn{5}{c}{$U(Y, A) = 93.1$} && \multicolumn{5}{c}{$U(Y, A) = 20.9$}  \\
  
  \cmidrule{3-7} \cmidrule{9-13} \cmidrule{15-19}
  
  && FTU & Supp$_l$ & Supp$_h$ & CDP & DP && FTU & Supp$_l$ & Supp$_h$ & CDP & DP && FTU & Supp$_l$ & Supp$_h$ & CDP & DP\\[10pt]

  $U(\hat{Y}, A)$ && 4.3 & 0.5 & 1.0 & 3.5 & 0.3 && 68.3 & 0.0 & 9.6 & 32.9 & 0.0 && 23.0 & 0.0 & 3.9 & 18.5 & 0.4\\
  ROC AUC && 88.4 & 73.4 & 76.7 & 74.9 & 73.4 && 98.4 & 50.4 & 75.9 & 75.7 & 50.5 && 95.2 & 52.5 & 81.7 & 70.1 & 53.6\\
  Flip && 100.0 & 100.0 & 100.0 & 89.0 & 77.8 && 100.0 & 100.0 & 100.0 & 96.8 & 92.2 && 100.0 & 100.0 & 100.0 & 79.1 & 55.8\\
  DP-ratio && 32.8 & 73.5 & 63.3 & 37.9 & 83.7 && 5.8 & 99.4 & 47.5 & 48.8 & 99.8 && 28.5 & 99.4 & 62.7 & 53.4 & 96.7\\
  
  \bottomrule
\end{tabular}
}
\end{table*}

The experiments are designed as follows: We assess a measure of pure group fairness, one of pure individual fairness, a measure of performance, and the amount of information embedded in the outcomes of different algorithms built with different strategies targeted to mitigate specific types of biases.
Namely, we test the behaviour of four different algorithms:
\begin{enumerate}
    \item an algorithm trained on non-sensitive features only (namely, on samples from $(X, Y)$), i.e. with a FTU approach;  
    \item an algorithm trained on features that have low correlation with $A$, i.e. with a Suppression approach; in particular we use 2 different correlation thresholds for each dataset in order to obtain 2 variants of the Suppression approach: one \emph{low} threshold allowing in the training set only variables with Pearson correlation lower than 5\% (both $X_1$ and $X_3$ are removed from the synthetic dataset, while 6 over 14 are removed from the Adult dataset), and a \emph{high} threshold allowing for more variables in the training set, namely with Pearson correlation up to 10\% for the Adult dataset (removing 3 over 14 variables) and up to 66\% for the synthetic experiments (removing only $X_1$);  
    \item an algorithm trained on the whole dataset (namely, on samples from $(X, A, Y)$), but then imposing acceptance rate parity for $A=0, 1$ given the value of other variables, i.e. imposing (a form of) CDP;
    \item an algorithm trained on the whole dataset (namely, on samples from $(X, A, Y)$), but then imposing acceptance rate parity for $A=0, 1$ unconditionally, i.e. imposing DP.
\end{enumerate}
We implement classifiers as random forests, while CDP and DP are imposed via a post-processing approach~\citep{hardt2016equality} by employing the \texttt{Fairlearn} python library~\citep{bird2020fairlearn}.
CDP is computed conditioned on $X_3$ in the synthetic setting while we condition on marital status in the Adult experiment.
The measures that we track are:
\begin{itemize}[-]
    \item Symmetric Uncertainty~\citep{witten2002data} $U(A, B) = \tfrac{2MI(A, B)}{H(A) + H(B)}$ --- which is the normalization of Mutual Information $MI$ with respect to the mean entropy $\tfrac{H(A) + H(B)}{2}$ --- to assess the interdependence between $\hat{Y}$ and $A$;
    \item Demographic Parity ratio of un-favoured over favoured group with respect to $A$ as a measure of pure group fairness (DP-ratio = 1 meaning perfect parity);
    \item To measure individual fairness, we build an auxiliary dataset where we flip the value of $A \rightarrow |1 - A|$, then compute the prediction $\hat{y}'_i$ of each algorithm on this ``fake'' data, and then compute the following metric
    \begin{equation}
        \text{Flip} = \frac{1}{n} \sum_{i=1}^n \lvert\hat{y}_i - \hat{y}'_i\rvert;
    \end{equation}
    \item Area Under the ROC Curve (ROC AUC) as a classification performance metric.
\end{itemize}

Table~\ref{tab:experiment} summarizes the results of the experiments on the 3 datastes, that, in general, corroborate the qualitative view outlined in this section. FTU is the approach which exploits the highest fraction of the information of $A$ contained in the data and reflecting in $Y$, DP and Suppression with low threshold have the opposite behaviour, with CDP and ``mild'' Suppression (Supp$_h$) in-between. A similar pattern results for performance, where full Suppression (Supp$_l$) ranks at the bottom together with DP-compliant models, while FTU, employing all the information in $X$, is always the best performing model. In terms of Flip metric, FTU and Suppression methods are by-design fully individual in this respect, since they don't explicitly employ $A$, while DP-compliant approaches negatively impact this metric, as expected. As for DP-ratio, FTU reaches very low values, in particular when the Mutual Information of $Y$ and $A$ is high, as expected, since FTU exploits information in $X$ to actually recover the $Y$ dependence on $A$ (indeed, as hinted above, FTU has the highest values of $U(\hat{Y}, A)$).  The results of the 2 variants of Suppression show that this approach behaves more like FTU or DP-compliant depending on the variables that we include in the training set.    
Notice that when Mutual Information of $Y$ and $A$ is high, then DP-compliant methods as well as full Suppression perform very poorly, since they try to impose the condition $\hat{Y} \indep A$, which is of course hardly compatible with an high performance.

\end{document}